\documentclass[11pt]{article}
\usepackage[table]{xcolor}
\usepackage[preprint]{acl}
\usepackage{times}
\usepackage{latexsym}
\usepackage[utf8]{inputenc}
\usepackage[T1]{fontenc}
\usepackage{microtype}
\usepackage{amsmath,amssymb,amsfonts}
\usepackage{booktabs,tabularx,array,multirow}
\usepackage{graphicx}
\definecolor{oursgray}{gray}{0.94}
\newcommand{\singlefigurewidth}{0.80\linewidth}

\usepackage{algorithm,algpseudocode}
\usepackage{enumitem}
\usepackage{stfloats}
\usepackage[above,below]{placeins}
\usepackage{url}
\usepackage{hyperref}
\makeatletter\let\tableinput\@@input\makeatother
\title{PathCal: Marker-Competition Decoding Control for Reasoning Models}
\author{%
Lingyu Jiang$^1$, Zirui Li$^1$, Lingyu Xu$^3$, Shuo Xing$^2$, Peiran Li$^2$, Tsubasa Takahashi$^1$,\\
\bf Dengzhe Hou$^1$, Zhengzhong Tu$^2$, Kazunori Yamada$^{1\star}$, Fangzhou Lin$^{1,2,3\dagger\star}$
\\[2pt]
$^1$Tohoku University\quad $^2$Texas A\&M University\quad $^3$Worcester Polytechnic Institute
\\[2pt]
\small $^\dagger$Project Lead.\quad $^\star$Corresponding Authors: \texttt{yamada@tohoku.ac.jp}, \texttt{fanghzou1@tamu.edu}.
}
\begin{document}
\raggedbottom
\maketitle
\begin{abstract}
Large Reasoning Language Models (LRMs) tackle complex tasks through test-time scaling, generating long-form Chain-of-Thought (CoT) trajectories during inference. These trajectories contain reflection markers such as \emph{wait}, \emph{but}, and \emph{alternatively}, accompanying hesitation, revision, and alternative explorations. Recent test-time control methods use these markers to steer reasoning, but treating them as one coarse-grained category can obscure their distinct intervention effects. We study how marker identity and reasoning context shape these effects. Type-wise suppression produces different accuracy--length tradeoffs, while fixed-prefix interventions show that the relative benefit of continuation and revision varies across prefixes and tasks. Motivated by these findings, we introduce \textsc{PathCal}, a training-free controller that adjusts the strength of a fixed continuation bias using instantaneous competition among continuation, revision, and alternative-opening markers. It applies finite logit adjustments without additional model forward passes. An exact log-odds identity and frozen-prefix probes establish its local directional effect. Experiments show that PathCal improves the accuracy--length tradeoff in many mathematical reasoning settings and also reduces generation length in coding and multilingual mathematics.
\end{abstract}

\section{Introduction}
Reasoning models such as DeepSeek-R1, QwQ, and OpenAI o3 tackle complex problems by extending computation at inference time \citep{guo2025deepseek,qwen2025qwq,openaio3}. Longer reasoning traces allow a model to explore solutions, check intermediate steps, and revise its answer, but also leave room for repetition and unnecessary detours. This raises a central question for test-time control: how can a fixed model use fewer generated tokens while retaining the benefits of extended reasoning? Discourse markers offer an accessible intervention interface. Expressions such as \emph{so}, \emph{but}, and \emph{alternatively} can introduce continuation, revision, or a new line of reasoning, and their scores are available during decoding.

\begin{figure*}[!tbp]
\centering
\includegraphics[width=\textwidth]{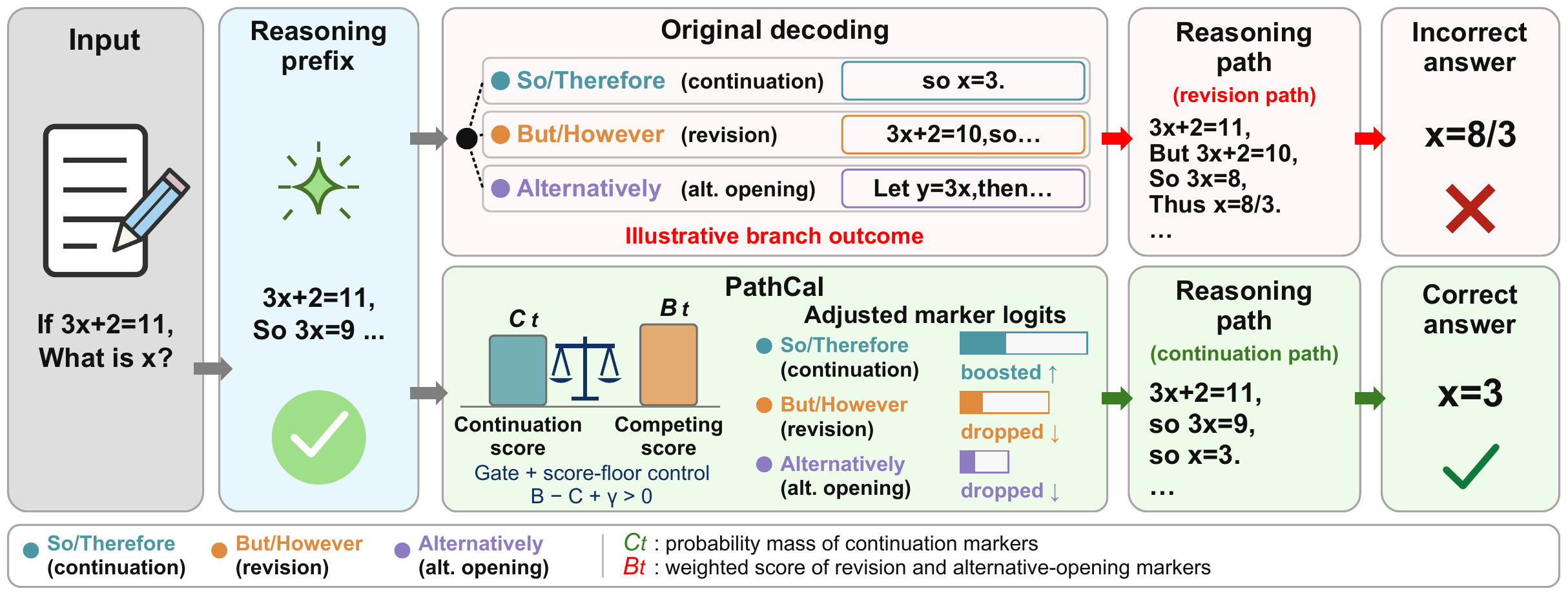}
\caption{PathCal modulates a fixed continuation bias using marker competition. The reasoning branches illustrate possible outcomes.}
\label{fig:pathcal_overview}
\end{figure*}

Repeated sampling allocates computation across candidate solutions \citep{brown2024largelanguagemonkeysscaling}, while Self-Refine uses iterative feedback \citep{madaan2023selfrefine}. Work on self-improving reasoners examines verification and backtracking behaviors \citep{gandhi2025cognitivebehaviorsenableselfimproving}. Thinking-token studies report both information-rich transitions \citep{qian2025demystifyingreasoningdynamicsmutual} and inefficient overthinking under constrained budgets \citep{ding2025thinkingtokenshelptrap}. These findings motivate studying how marker identity and the surrounding reasoning prefix shape an intervention's effect.

Existing approaches already manipulate reasoning length and reflection markers. Thought Switching Penalty (TIP) applies a penalty within a thought-dependent window \citep{wang2025tip}, and CyclicReflex modulates reflection tokens by generation position \citep{fan2026cyclicreflex}. We investigate category-specific control driven by competition in the current next-token distribution, complementing thought-window and position-based intervention schedules.

Our starting observation is that marker identity matters. Suppressing all selected markers shortens reasoning at an accuracy cost, whereas suppressing individual marker types produces different accuracy--length tradeoffs. Moreover, fixed-prefix interventions with \emph{So} and \emph{But} yield point-estimate contrasts that vary in magnitude and direction across prefix-value bins and evaluation settings. These findings motivate distinguishing marker groups and considering the context of intervention.

We translate this motivation into \textsc{PathCal} (Figure~\ref{fig:pathcal_overview}), a controller driven by competition in the current next-token distribution. It groups markers into continuation, revision, and alternative opening, then uses their weighted scores to modulate finite logit adjustments. The direction is fixed: continuation is encouraged, while competing openings are suppressed. What adapts is the intervention strength. This design favors carrying a line of reasoning forward, with a gate and score-gap factor regulating the bias at each step. It requires no extra model forward pass.

The resulting controller improves accuracy in 19 of 20 model--benchmark pairs and shortens generations in 18 of 20 in our single-seed main comparison. Five-seed coding and multilingual mathematics comparisons also show shorter generations, with clearer task-score gains in coding. An exact log-odds identity and frozen-prefix probes explain the controller's local directional effect; ablations then examine how activation, direction, and vocabulary shape downstream performance. Independent development comparisons show that the accuracy benefit varies across settings.

Our contributions are:
\begin{itemize}[leftmargin=*,label=\textbullet,itemsep=2pt,topsep=3pt,parsep=0pt]
\item \textbf{Marker-dependent reasoning interventions.} We show that type-wise suppression changes the accuracy--length tradeoff, and that the relative success of forced continuation and revision depends on the prefix and task. These findings motivate category-specific control that accounts for the context of intervention.
\item \textbf{Marker-competition decoding control.} We introduce \textsc{PathCal}, which adapts the strength of a fixed continuation bias to current marker scores using finite, non-masking logit adjustments. We derive its exact local log-odds effect and verify the directional response at frozen prefixes, without additional model forward passes.
\item \textbf{Reasoning performance and design insights.} We evaluate four models on five main mathematical benchmarks, complemented by TheoremQA, coding, and multilingual mathematics. Component ablations and direction and vocabulary studies explain the accuracy--length tradeoffs and identify settings in which stronger compression does not improve accuracy.
\end{itemize}

\FloatBarrier
\section{Related Work}
\label{sec:related}

\paragraph{Efficient reasoning and compression.}
Chain-of-thought prompting and reasoning post-training can improve reasoning through longer traces \citep{wei2022chain,kojima2023largelanguagemodelszeroshot,guo2025deepseek}. Approaches to reducing this cost include concise prompting \citep{aytes2025sketchofthoughtefficientllmreasoning,xu2025chaindraftthinkingfaster}, token or trace compression \citep{xia2025tokenskip,r1compress2025,qiao2025concise}, and reasoning-path search \citep{xu2025astarthought}. PathCal controls generated-token usage within a single rollout through marker-level decoding adjustments.

\paragraph{Test-time scaling and adaptive computation.}
Verifier-based selection, self-consistency, and tree search allocate compute across candidate solutions \citep{lightman2024lets,wang2022selfconsistency,yao2023treethoughtsdeliberateproblem,feng2024alphazeroliketreesearchguidelarge}. Adaptive methods regulate budgets through length constraints, difficulty, reflection preferences, or confidence-based stopping \citep{snell2025scaling,muennighoff2025s1,aggarwal2025l1controllinglongreasoning,huang2025adactrladaptivecontrollablereasoning,wan2025adapthinkadaptivethinkingpreferences,fu2025efficientlyscalingllmreasoning,yang2025dynamicearlyexitreasoning}. Analyses of test-time scaling also examine when longer traces help \citep{su2025underthinkingoverthinkingempiricalstudy,zeng2025revisitingtesttimescalingo1like,yang2025thinkingoptimalscalingtesttimecompute,wu2025itssimpleanalysissimple}. PathCal complements these budget-level approaches by adjusting local marker preferences.

\paragraph{Decoding and representation interventions.}
Contrastive decoding modifies token distributions using reference models \citep{li-etal-2023-contrastive,tang2026thinkingsubtractionconfidencedrivencontrastive}, while inference-time interventions modify hidden activations \citep{li2023inferencetime}. PathCal adjusts selected marker logits from the same forward pass. Analyses of uncertainty, reflection, backtracking, and reasoning-step influence motivate studying intervention targets \citep{galichin2025icoveredbaseshere,ward2025reasoningfinetuningrepurposeslatentrepresentations,bogdan2025thoughtanchorsllmreasoning}; controlled working-memory tasks offer a complementary behavioral perspective \citep{hou2026wmf}.

\paragraph{Marker-based reasoning control.}
TIP penalizes thought-switching markers within post-boundary windows \citep{wang2025tip}, while CyclicReflex applies position-based cyclical modulation \citep{fan2026cyclicreflex}. PathCal distinguishes continuation, revision, and alternative-opening groups and uses their weighted next-token scores to modulate a fixed continuation bias. Its distinguishing choice is the combination of category-specific updates and instantaneous marker competition as the control signal. Section~\ref{sec:experimental_setup} identifies the evaluated baseline variants.

\section{Understanding Marker-Dependent Reasoning Interventions}
\label{sec:reflection_markers}
Marker-based control depends on both the expression being changed and the reasoning context in which it appears. Type-wise suppression reveals different accuracy--length responses across marker types; fixed-prefix interventions show how the relative benefit of continuation and revision varies across prefixes. Together, they motivate distinguishing marker groups and considering intervention context.

\subsection{Marker Identity Changes the Tradeoff}
\label{subsec:typewise_suppression}
Treating reflection-related markers as a single control target overlooks the different roles their occurrences can play. We test the consequences by comparing standard decoding (\textsc{Original}), suppression of all selected markers (\textsc{SuppressAll}), and suppression of individual types (\textsc{SuppressOnly}), including \emph{wait}, \emph{but}, and \emph{however}, on Qwen-7B AIME2025.

For a marker group $\mathcal G_g$, the intervention during reasoning is
\begin{equation}
\widetilde z_{t,v}^{(g)}=z_{t,v}-\lambda\,\mathbf1[v\in\mathcal G_g],
\qquad \lambda=5.
\label{eq:diagnostic_suppression}
\end{equation}
It makes selected markers less likely without removing them from the vocabulary, and stops at the reasoning-end token.

\begin{figure}[!htbp]
\centering
\includegraphics[width=\singlefigurewidth]{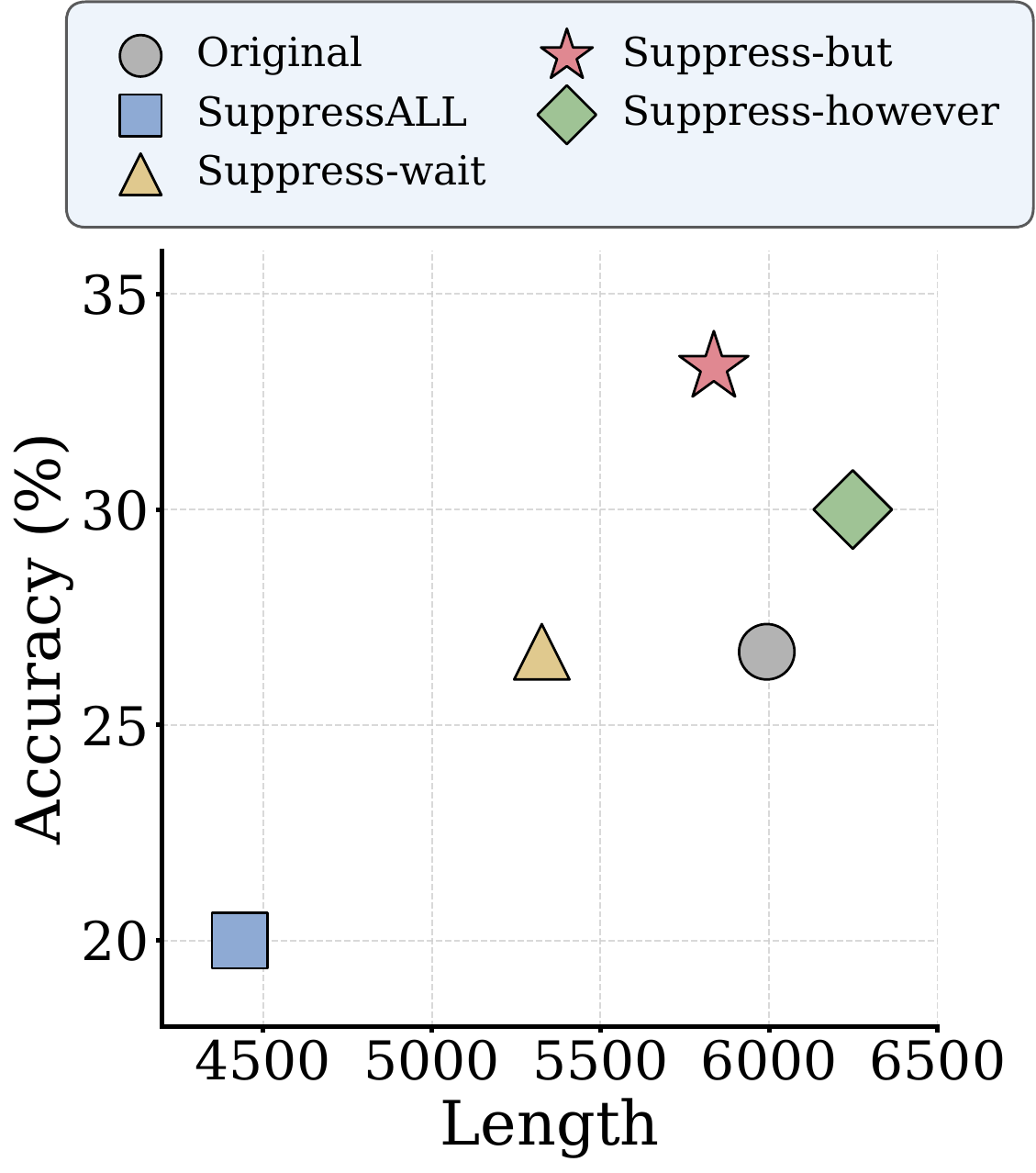}
\caption{Type-wise marker suppression on AIME2025 using Qwen-7B.}
\label{fig:suppression_tradeoff}
\label{fig:suppression}
\end{figure}

Figure~\ref{fig:suppression_tradeoff} shows that blanket suppression achieves the greatest shortening at an accuracy cost. Selective suppression follows a different pattern: suppressing \emph{but} raises accuracy without severe shortening, suppressing \emph{wait} yields moderate shortening with similar accuracy, and suppressing \emph{however} raises accuracy while lengthening the trace. Thus, the effect of marker intervention cannot be reduced to how much reasoning it removes. Which marker is suppressed changes the resulting tradeoff.

\subsection{Marker Effects Depend on the Prefix}
\label{subsec:counterfactual_branch}
\begin{figure*}[!tbp]
\centering
\includegraphics[width=\textwidth]{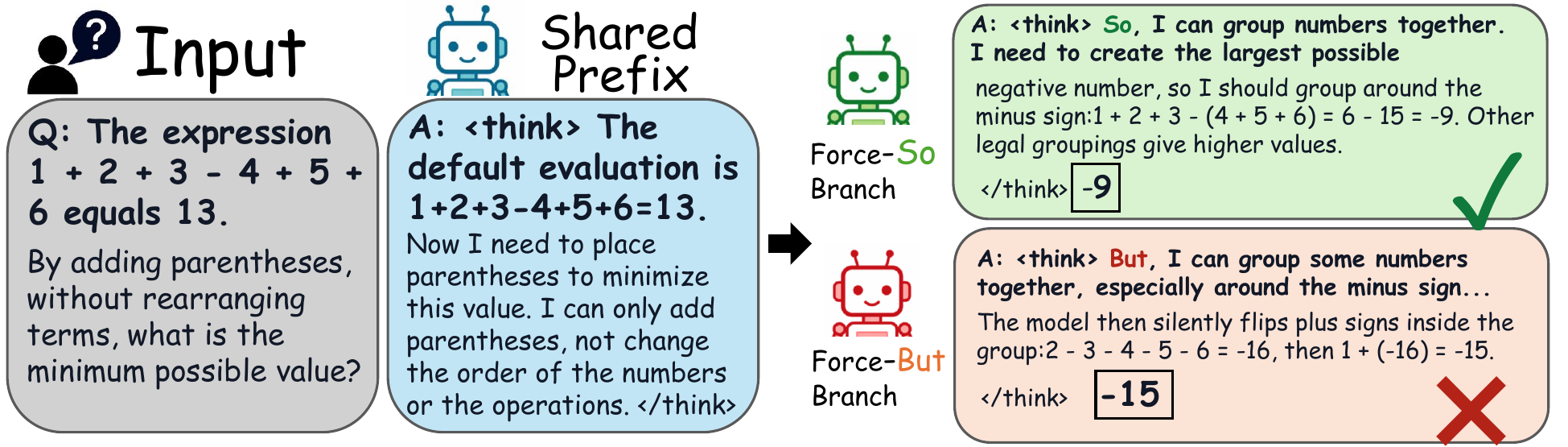}
\caption{Fixed-prefix branch intervention: forcing \emph{So} or \emph{But} after the same input and prefix can lead to different continuations.}
\label{fig:branch_example}
\end{figure*}

The suppression study changes marker preferences throughout generation. To understand the local effect of marker choice, we instead hold the input and reasoning prefix fixed and change only the next marker. We compare \emph{So}, which opens a continuation, with \emph{But}, which opens a revision, then sample the remaining reasoning normally. Figure~\ref{fig:branch_example} illustrates how the two openings can lead to different continuations from the same prefix.

For prefix $s$ and candidate marker $m$, define
\begin{equation}
V(s,m)=\mathbb E_{\tau\sim p_\theta(\cdot\mid s,m)}
\left[\mathbf1[\mathrm{Correct}(\tau)]\right],
\label{eq:diagnostic_prefix_value}
\end{equation}
the probability of reaching a correct answer after forcing $m$. Unforced continuations estimate the ordinary prefix value $\widehat V(s)$, which groups prefixes into low-, middle-, and high-value bins. This groups reasoning contexts by sampled continuation success rather than by problem-level difficulty. We study Qwen-7B on MATH500, AIME2024, and AIME2025; sampling details appear in Appendix~\ref{app:diagnostic_details}.

\begin{table}[!htbp]
\centering\small
\setlength{\tabcolsep}{4pt}
\renewcommand{\arraystretch}{1.0}
\begin{tabular*}{\linewidth}{@{}@{\extracolsep{\fill}}ccccc@{}}
\toprule
Dataset / cap & Bin & $V_{\mathrm{So}}$ & $V_{\mathrm{But}}$ & $\Delta$ \\
\midrule
\multirow{3}{*}{\shortstack{MATH500\\8k}} & low & 7.1 & 10.0 & -3.0 \\
 & mid & 57.5 & 50.3 & \textbf{+7.1} \\
 & high & 97.6 & 97.7 & -0.1 \\
\midrule
\multirow{3}{*}{\shortstack{AIME24\\16k}} & low & 4.8 & 10.5 & -5.6 \\
 & mid & 58.8 & 52.5 & \textbf{+6.2} \\
 & high & 89.7 & 89.1 & +0.6 \\
\midrule
\multirow{3}{*}{\shortstack{AIME25\\16k}} & low & 8.2 & 7.1 & +1.1 \\
 & mid & 36.8 & 47.1 & \textbf{-10.3} \\
 & high & 95.4 & 96.3 & -0.9 \\
\midrule
\multirow{3}{*}{\shortstack{AIME24\\8k}} & low & 9.0 & 11.7 & -2.7 \\
 & mid & 41.7 & 56.2 & \textbf{-14.6} \\
 & high & 90.3 & 88.7 & +1.6 \\
\bottomrule
\end{tabular*}
\caption{Fixed-prefix interventions. Values are percentages; $\Delta=V_{\mathrm{So}}-V_{\mathrm{But}}$ is in percentage points. Bold marks the middle bin. Budgets are continuation caps; AIME24 prefix sets differ across runs. Sampling details and problem-cluster intervals appear in Appendix~\ref{app:diagnostic_details}.}
\label{tab:branch_intervention}
\end{table}

Table~\ref{tab:branch_intervention} shows the largest point-estimate contrasts in the middle-value bins, where ordinary continuations include both successes and failures. The observed So--But contrasts vary in magnitude and direction across prefix-value bins and evaluation settings, suggesting heterogeneous intervention effects. The high-value bins show smaller differences, alongside their more limited room for improvement. Appendix~\ref{app:diagnostic_details} reports the full results and problem-cluster confidence intervals.

\paragraph{From marker effects to decoding control.}
PathCal uses next-token marker scores as an immediately available control signal, letting their competition determine how strongly to favor continuation. This design is motivated by the observed heterogeneity; the diagnostics do not select its gate or establish an optimal intervention direction. Section~\ref{sec:experiments} studies the resulting tradeoff.

\FloatBarrier
\section{Marker-Competition Decoding Control}
\label{sec:pathcal}

\textsc{PathCal} adapts intervention strength to instantaneous marker competition while keeping its direction fixed. It favors continuation through finite logit adjustments, using a gate and score-gap factor to regulate the bias at each step. This design encourages carrying the current line of reasoning forward and uses marker scores as its online control signal. Reverse control is a diagnostic comparison, not a second direction selected during PathCal decoding.

\subsection{Marker Scores}
\label{subsec:pathcal_marker_scores}

Let $h_t$ denote the input and generated prefix at step $t$, and let
$\ell_t(v)$ be the logits supplied to the controller for that history. We compute
$p_t=\operatorname{softmax}(\ell_t)$ before applying its adjustment.
Three disjoint token-ID sets represent operational marker groups:
continuation $\mathcal M_C$ (e.g., ``So''), revision $\mathcal M_R$
(e.g., ``But''), and alternative opening $\mathcal M_A$
(e.g., ``Alternatively''). Group membership is determined by token identity. The manually specified surface forms
are resolved to single-token IDs separately for each tokenizer
(Appendix~\ref{app:markers}).

The controller aggregates their probabilities as
\begin{equation}
\begin{aligned}
C_t&=\sum_{v\in\mathcal M_C}p_t(v), &
A_t&=\sum_{v\in\mathcal M_A}p_t(v),\\
R_t&=\sum_{v\in\mathcal M_R}w_vp_t(v),
\end{aligned}
\label{eq:pathcal_scores}
\end{equation}
where $w_v\geq1$ weights selected revision markers, and forms the
competing-marker score
\begin{equation}
B_t=R_t+\lambda_A A_t.
\label{eq:pathcal_competing_score}
\end{equation}
The weights make $R_t$, $B_t$, and $C_t+B_t$ marker scores rather than unweighted probability masses.

\subsection{Competition Gate and Logit Adjustment}
\label{subsec:pathcal_rule}

If $C_t+B_t<\rho$, the controller leaves the logits unchanged.
Otherwise, it computes
\begin{equation}
g_t=\frac{4C_tB_t}{(C_t+B_t)^2+\epsilon},
\label{eq:pathcal_gate}
\end{equation}
with $\epsilon>0$. This bounded gate favors balanced marker scores
while attenuating low total scores. A second factor sets the adjustment
strength according to the competing-marker gap:
\begin{equation}
\begin{aligned}
\alpha_t={}&\mathbf1[C_t+B_t\geq\rho]\,
\alpha_{\mathrm{base}}g_t\\[-1mm]
&\times\min\!\left\{\frac{[B_t-C_t+\gamma]_+}{\tau},1\right\},
\end{aligned}
\label{eq:pathcal_alpha}
\end{equation}
where $[a]_+=\max(a,0)$, $\gamma$ is a margin, and $\tau>0$
sets gap saturation. The gap factor is positive when $B_t>C_t-\gamma$,
so a positive margin permits intervention even when $B_t$ is slightly
below $C_t$. Strength need not grow monotonically with $B_t$,
because highly unbalanced scores reduce the gate.

For positive coefficients $\beta_C,\beta_R,\beta_A$, the update is
\begin{equation}
\tilde\ell_t(v)=\ell_t(v)+\alpha_t
\begin{cases}
\beta_C,&v\in\mathcal M_C,\\
-\beta_Rw_v,&v\in\mathcal M_R,\\
-\beta_A,&v\in\mathcal M_A,\\
0,&\text{otherwise}.
\end{cases}
\label{eq:pathcal_shift}
\end{equation}
The base sampler then uses the adjusted logits. This non-masking update increases
continuation logits and decreases revision and alternative-opening
logits by finite amounts; finite does not imply small. Other logits remain
unchanged, although normalization can change their probabilities.
The gate controls the strength of this category-specific adjustment at each decoding step.

\paragraph{Local control property.}
\label{subsec:pathcal_property}
Let $q_t=\operatorname{softmax}(\tilde\ell_t)$. For continuation marker
$c$ and revision marker $r$ with finite logits,
\begin{equation}
\begin{aligned}
\log\frac{q_t(c)}{q_t(r)}-\log\frac{p_t(c)}{p_t(r)}
&=(\beta_C+\beta_Rw_r)\alpha_t.
\end{aligned}
\label{eq:pathcal_revision_odds}
\end{equation}
For an alternative-opening marker $a$, the corresponding increment is
$(\beta_C+\beta_A)\alpha_t$. These identities precisely characterize
the local change in relative marker odds. They concern the distributions
at the controller, before any subsequent temperature scaling or
truncation.
Appendix~\ref{app:logodds} provides the proof and sampler conditions.

\paragraph{Decoding integration.}
The decoding integration uses vLLM \citep{kwon2023efficientmemorymanagementlarge}. It assumes that generation begins in the reasoning
region. Intervention begins after a 100-token warmup and ceases once the
generated sequence contains \texttt{</think>}. It adds no model forward
pass, verifier, or sampled candidate. The canonical implementation uses
a full-vocabulary softmax followed by marker-score computation and
logit edits. Algorithm~\ref{alg:pathcal_decoding} summarizes the decoding procedure.
The marker inventory and canonical defaults appear in
Appendices~\ref{app:markers} and~\ref{app:algorithm}, with computational
details in Appendix~\ref{app:cost}.

\begin{algorithm}[!t]
\caption{\textsc{PathCal} decoding}
\label{alg:pathcal_decoding}
\small
\begin{algorithmic}[1]
\Require Base model and sampler; prompt initialized for reasoning;
disjoint marker sets $\mathcal M_C,\mathcal M_R,\mathcal M_A$;
revision weights $w_v$; canonical parameters
\State $z\gets()$
\For{$t=1,\ldots,T_{\max}$}
    \State Obtain next-token logits $\ell_t$ from the base model
    \State $\tilde\ell_t\gets\ell_t$
    \If{$|z|\geq\mathrm{minp}$ and $\texttt{</think>}\notin z$}
        \State $p_t\gets\operatorname{softmax}(\ell_t)$
        \State Compute $C_t,R_t,A_t,B_t$ using
        Eqs.~\eqref{eq:pathcal_scores}--\eqref{eq:pathcal_competing_score}
        \If{$C_t+B_t\geq\rho$}
            \State Compute $g_t$ and $\alpha_t$ using
            Eqs.~\eqref{eq:pathcal_gate}--\eqref{eq:pathcal_alpha}
            \State Update $\tilde\ell_t$ by Eq.~\eqref{eq:pathcal_shift}
        \EndIf
    \EndIf
    \State Sample $z_t$ from $\tilde\ell_t$ with the base sampler
    \State Append $z_t$ to $z$
    \If{$z_t$ is an end-of-sequence token}
        \State \textbf{break}
    \EndIf
\EndFor
\State \Return $z$
\end{algorithmic}
\end{algorithm}

\section{Experiments}
\label{sec:experiments}
We evaluate whether marker-competition control can improve reasoning accuracy while reducing generated-token usage. We then examine how far the length reduction extends beyond the main mathematical benchmarks, and how activation, direction, and marker vocabulary influence the resulting tradeoff.

\subsection{Experimental Setup}
\label{sec:experimental_setup}
\noindent\textbf{Models.}
We evaluate \textsc{PathCal} on four open-source reasoning models spanning scale, backbone architecture, and training pipeline.
\textbf{DeepSeek-R1-Distill-Qwen-7B} and \textbf{DeepSeek-R1-Distill-Qwen-14B}~\citep{guo2025deepseek} provide two scales within the Qwen-based~\citep{qwen2025qwen25technicalreport} DeepSeek-R1 distilled family.
\textbf{DeepSeek-R1-Distill-Llama-8B}~\citep{guo2025deepseek} tests transfer to a different Llama-based backbone~\citep{grattafiori2024llama3herdmodels}.
\textbf{QwQ-32B}~\citep{qwen2025qwq} provides a non-DeepSeek-distilled reasoning model.
Together, this suite tests marker-level calibration across scales, backbones, and distillation pipelines.

\noindent\textbf{Datasets.}
Our evaluation covers mathematical reasoning, coding, and multilingual mathematics. The main comparison in Table~\ref{tab:main_grouped} uses five mathematical benchmarks.
\textbf{GSM8K}~\citep{cobbe2021gsm8k,vendrow2025largelanguagemodelbenchmarks} covers grade-school arithmetic reasoning.
\textbf{MATH500}~\citep{hendrycks2021math,lightman2024lets}, \textbf{AMC2023}~\citep{aimo2024amc}, \textbf{AIME2024}, and \textbf{AIME2025}~\citep{balunović2026matharenaevaluatingllmsuncontaminated} cover competition-style mathematics.
A separate comparison on \textbf{TheoremQA}~\citep{chen-etal-2023-theoremqa} examines theorem-based reasoning across science and mathematics (Figure~\ref{fig:theoremqa_tradeoff}).
We additionally evaluate code generation on \textbf{LiveCodeBench}~\citep{jain2024livecodebench} and multilingual mathematics on \textbf{MGSM}~\citep{shi2023mgsm}, using Qwen-7B and Llama-8B with the same marker inventory and default parameters (Table~\ref{tab:domain_summary}). LiveCodeBench uses a 279-problem subset; MGSM uses 50 base problems translated into 11 aligned language versions. Dataset details appear in Table~\ref{tab:app_datasets}.

\noindent\textbf{Baselines.}
We compare \textsc{PathCal} with four training-free test-time baselines.
\textbf{Original} uses standard decoding without intervention.
\textbf{TIP}~\citep{wang2025tip} denotes the reflection-marker penalty variant used in our comparison.
\textbf{CyclicReflex}~\citep{fan2026cyclicreflex} cyclically modulates reflection-marker logits.
The \textbf{s1-style} baseline~\citep{muennighoff2025s1}, labeled S1 in Table~\ref{tab:main_grouped}, prompts additional reasoning through reflection cues.
Together, they cover standard decoding, reflection-marker control, and prompted continuation, while \textsc{PathCal} uses category-specific marker scores to adjust decoding.

\noindent\textbf{Evaluation protocol.}
We report \textbf{final-answer accuracy} and \textbf{average generated tokens} under single-sample decoding; LiveCodeBench uses \textbf{pass@1} as its task score.
Decoding uses vLLM~\citep{kwon2023efficientmemorymanagementlarge}.
The main comparison uses seed 42; coding and multilingual results average five seeds. Appendix~\ref{app:evidence} specifies the protocols for these evaluations, the independent development comparisons, and the prefix probes. Controller defaults appear in Table~\ref{tab:app_pathcal_hparams}. Our cost metric is generated-token usage; Appendix~\ref{app:cost} analyzes the additional per-step computation.

\subsection{Reasoning Accuracy and Generation Length}
\label{sec:task_results}
\begin{table*}[t]
\centering
\caption{
 Main results on five mathematical reasoning benchmarks.
Acc. denotes accuracy (\%), and Len. denotes generated tokens.
Best and second-best values within each model block are shown in \textbf{bold} and \underline{underlined}, respectively.
\textsc{PathCal} is highlighted in gray.}
\label{tab:main_grouped}
\setlength{\tabcolsep}{3.2pt}
\renewcommand{\arraystretch}{0.8}
\scriptsize
\resizebox{0.9\textwidth}{!}{%
\begin{tabular}{l |cc|cc|cc|cc|cc}
\toprule
\multirow{2}{*}{Method}
& \multicolumn{2}{c}{MATH500}
& \multicolumn{2}{c}{AIME24}
& \multicolumn{2}{c}{AIME25}
& \multicolumn{2}{c}{AMC23}
& \multicolumn{2}{c}{GSM8K} \\
\cmidrule(lr){2-3}
\cmidrule(lr){4-5}
\cmidrule(lr){6-7}
\cmidrule(lr){8-9}
\cmidrule(lr){10-11}
& Acc.\,$\uparrow$ & Len.\,$\downarrow$
& Acc.\,$\uparrow$ & Len.\,$\downarrow$
& Acc.\,$\uparrow$ & Len.\,$\downarrow$
& Acc.\,$\uparrow$ & Len.\,$\downarrow$
& Acc.\,$\uparrow$ & Len.\,$\downarrow$ \\
\midrule

\multicolumn{11}{c}{DeepSeek-R1-Distill-Qwen-7B} \\
\midrule
Original      & 85.6 & 1423 & 33.3 & 5678 & 26.7 & 5606 & 77.5 & \underline{2750} & 86.8 & 339 \\
TIP           & 86.2 & \underline{1406} & 36.7 & \textbf{4772} & 26.7 & 5609 & \textbf{82.5} & 2893 & 86.2 & \underline{323} \\
S1            & \underline{86.6} & 1790 & \underline{40.0} & 6154 & 26.7 & 6035 & \textbf{82.5} & 3335 & 86.4 & 404 \\
CyclicReflex  & 85.8 & 1459 & 36.7 & 5308 & \underline{30.0} & \underline{5567} & \underline{80.0} & 2806 & \underline{86.9} & 380 \\
\rowcolor{oursgray}
PathCal(Ours) & \textbf{87.4} & \textbf{1281} & \textbf{43.3} & \underline{4990} & \textbf{36.7} & \textbf{5051} & \textbf{82.5} & \textbf{2244} & \textbf{87.0} & \textbf{319} \\

\midrule
\multicolumn{11}{c}{DeepSeek-R1-Distill-Qwen-14B} \\
\midrule
Original      & 87.8 & \underline{1978} & \underline{46.7} & \underline{6009} & 33.3 & 6683 & 82.5 & 3733 & 93.3 & \underline{480} \\
TIP           & 88.2 & 2008 & 40.0 & 6268 & \underline{36.7} & 6696 & 82.5 & 3733 & 93.0 & 482 \\
S1            & \underline{88.6} & 2391 & 43.3 & 6665 & \textbf{40.0} & 7020 & 77.5 & 3980 & \textbf{93.9} & 607 \\
CyclicReflex  & 86.8 & 2059 & \underline{46.7} & 6059 & 30.0 & \underline{6677} & \underline{85.0} & \underline{3695} & 93.5 & 537 \\
\rowcolor{oursgray}
PathCal(Ours) & \textbf{91.0} & \textbf{1851} & \textbf{56.7} & \textbf{5566} & \textbf{40.0} & \textbf{6014} & \textbf{90.0} & \textbf{3166} & \underline{93.6} & \textbf{479} \\

\midrule
\multicolumn{11}{c}{DeepSeek-R1-Distill-Llama-8B} \\
\midrule
Original      & \underline{82.6} & 1988 & 30.0 & 6620 & 23.3 & 6960 & \underline{75.0} & 3937 & 83.2 & 480 \\
TIP           & 81.4 & \underline{1932} & 30.0 & \underline{6440} & \underline{26.7} & 6562 & \textbf{77.5} & \underline{3310} & \textbf{84.9} & \underline{459} \\
S1            & 80.4 & 2205 & 33.3 & 7019 & 23.3 & 6818 & 72.5 & 4291 & 79.3 & 495 \\
CyclicReflex  & 81.4 & 2038 & \underline{40.0} & 6518 & \underline{26.7} & \underline{6424} & \underline{75.0} & 3937 & 82.9 & 482 \\
\rowcolor{oursgray}
PathCal(Ours) & \textbf{83.2} & \textbf{1867} & \textbf{46.7} & \textbf{5715} & \textbf{30.0} & \textbf{6202} & \underline{75.0} & \textbf{3064} & \underline{84.6} & \textbf{438} \\

\midrule
\multicolumn{11}{c}{QwQ-32B} \\
\midrule
Original      & 92.8 & 4209 & 73.3 & 12886 & 66.7 & \underline{15792} & 95.0 & 7524 & 96.1 & 1593 \\
TIP           & \underline{93.8} & \underline{4155} & \underline{76.7} & \underline{12854} & 63.3 & 16344 & \textbf{100.0} & \textbf{6454} & \underline{96.4} & \underline{1494} \\
S1            & \textbf{94.6} & 4549 & 73.3 & 14575 & \textbf{76.7} & 16048 & \underline{97.5} & 7295 & \underline{96.4} & 1830 \\
CyclicReflex  & 93.2 & 4261 & 73.3 & \textbf{12848} & \textbf{76.7} & \textbf{14733} & \textbf{100.0} & 7076 & \textbf{96.7} & 1581 \\
\rowcolor{oursgray}
PathCal(Ours) & \underline{93.8} & \textbf{3867} & \textbf{83.3} & 13426 & \underline{70.0} & 15989 & \textbf{100.0} & \underline{6608} & \textbf{96.7} & \textbf{1430} \\
\bottomrule
\end{tabular}%
}
\end{table*}

\paragraph{Main mathematical reasoning results.}
PathCal improves the accuracy--length tradeoff relative to standard decoding across most of the main comparison (Table~\ref{tab:main_grouped}). Accuracy increases in 19 of 20 model--benchmark pairs and is unchanged in the remaining pair; generations are shorter in 18 of 20. Thus, the gains in this evaluation accompany reduced token usage in most settings, extending beyond length reduction alone.

The comparison also shows where the tradeoff changes. On QwQ AIME2025, CyclicReflex is both more accurate and shorter than PathCal. On both QwQ AIME sets, PathCal's higher accuracy relative to Original comes with longer generations. A local continuation preference therefore need not produce a shorter complete solution: its downstream effect depends on the model and task.

\begin{figure}[!t]
\centering
\includegraphics[width=\linewidth]{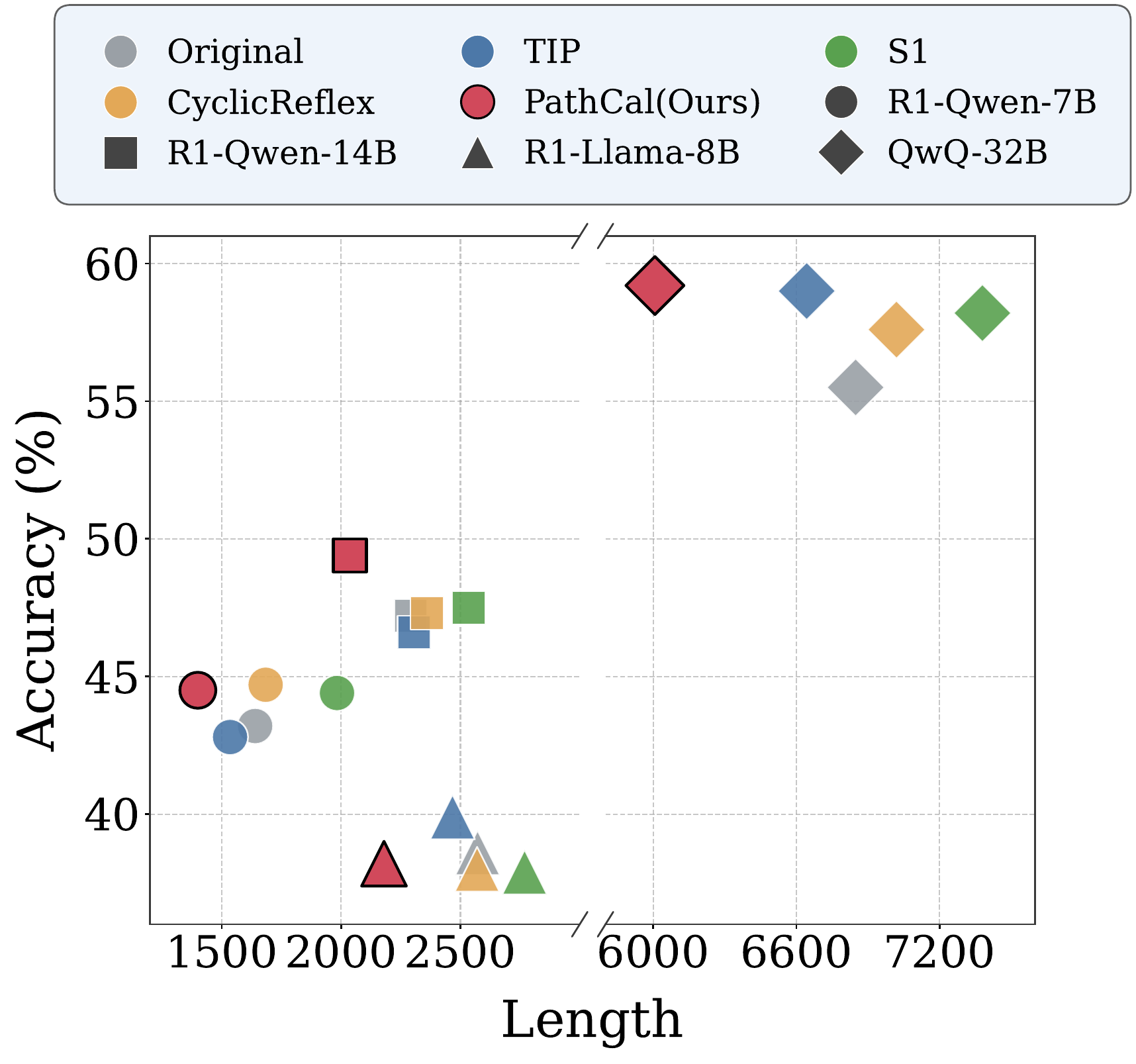}
\caption{Accuracy versus generation length on TheoremQA. Colors indicate methods and markers indicate model families.}
\label{fig:theoremqa_tradeoff}
\end{figure}

TheoremQA extends this comparison to theorem-based reasoning across science and mathematics (Figure~\ref{fig:theoremqa_tradeoff}), examining how the same marker control changes accuracy and length outside competition-style problems.

\paragraph{Coding and multilingual math.}
The additional tasks show two patterns (Table~\ref{tab:domain_summary}). On LiveCodeBench, both models have higher mean pass@1 and shorter generations across five seeds. On MGSM, generation length decreases, while mean accuracy is unchanged for Qwen-7B and increases marginally for Llama-8B.

\begin{table}[!t]
\centering\small
\setlength{\tabcolsep}{2pt}
\renewcommand{\arraystretch}{1.0}
\begin{tabular*}{\linewidth}{@{}@{\extracolsep{\fill}}lrrrr@{}}
\toprule
& \multicolumn{2}{c}{Score (\%) $\uparrow$} & \multicolumn{2}{c}{Tokens $\downarrow$} \\
\cmidrule(lr){2-3}\cmidrule(l){4-5}
Dataset & Original & PathCal & Original & PathCal \\
\midrule
\multicolumn{5}{l}{\textit{Qwen-7B}} \\
LCB & 30.97 & 32.11 & 6838.3 & 6616.7 \\
MGSM & 63.09 & 63.09 & 503.2 & 469.9 \\
\midrule
\multicolumn{5}{l}{\textit{Llama-8B}} \\
LCB & 26.59 & 29.32 & 7020.2 & 6802.8 \\
MGSM & 50.44 & 50.69 & 620.5 & 582.8 \\
\bottomrule
\end{tabular*}
\caption{Five-seed means on coding and multilingual mathematics. LCB denotes LiveCodeBench; Score is pass@1 for LCB and accuracy otherwise.}
\label{tab:domain_summary}
\end{table}

Token-length reductions thus extend to coding and multilingual mathematics, while task-score effects remain setting-dependent. MGSM primarily demonstrates shorter outputs with similar accuracy across aligned translations.

\subsection{Decoding Behavior and Design Analysis}
\label{sec:local_control}
\paragraph{Marker adjustments steer the next decision.}
We measure the controller's immediate effect by applying Forward or Reverse at the next decision of fixed AIME prefixes, varying the imposed strength and sampling the remaining continuation normally.

Table~\ref{tab:local_control} shows a strength-dependent response: Forward favors continuation, whereas Reverse favors revision or alternative openings. Frequency-matched ordinary tokens produce a much weaker response (Appendix Table~\ref{tab:ordinary_token_control}). Marker identity therefore provides a targeted handle for changing the next decision.

\begin{figure}[!tbp]
\centering
\includegraphics[width=\linewidth]{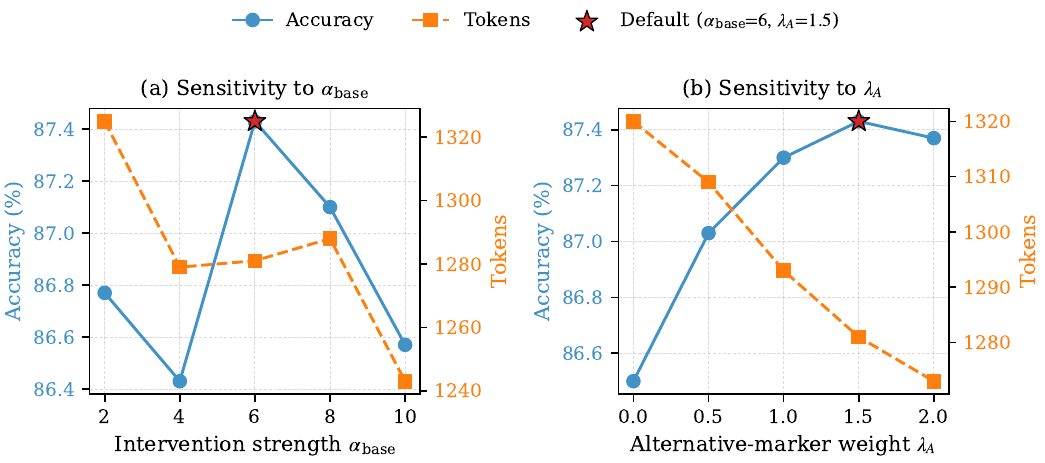}
\caption{MATH500 accuracy and generation length under varying base strength $\alpha_{\mathrm{base}}$ (left) and alternative-marker weight $\lambda_A$ (right).}
\label{fig:hyperparam_sensitivity}
\end{figure}
\paragraph{Stronger compression can cost accuracy.}
Removing the gate and marker-score floor together (NoGate/Floor) produces the shortest generations but returns mean accuracy to Original in the independent MATH500 study (Table~\ref{tab:historical_ablation_main}). The combined activation rule moderates this accuracy cost. Removing the alternative group (NoAltGroup), continuation boost, or revision suppression also lowers mean accuracy relative to Full.

Among 40 triggered prefixes in a hard-marker probe, continuation succeeds more often at 3, revision at 4, and 33 are sampled ties with four rollouts per direction (Appendix Table~\ref{tab:gate_alignment}). Thus, this probe does not show that activation selects prefixes favoring continuation.

\begin{table}[!t]
\centering\small
\setlength{\tabcolsep}{4pt}
\renewcommand{\arraystretch}{1.0}
\begin{tabular}{@{}lrr@{}}
\toprule
Condition & Cont. (\%) & Rev.+Alt. (\%) \\
\midrule
Natural & 61.04 & 25.21 \\
\midrule
Forward, $\alpha=1.5$ & 68.75 & 13.12 \\
Forward, $\alpha=3$ & 75.00 & 4.17 \\
Forward, $\alpha=6$ & 77.92 & 0.83 \\
\midrule
Reverse, $\alpha=1.5$ & 50.62 & 33.96 \\
Reverse, $\alpha=3$ & 38.54 & 42.50 \\
Reverse, $\alpha=6$ & 16.46 & 65.00 \\
\bottomrule
\end{tabular}
\caption{First-token category frequencies. Each condition has 480 generations from 120 frozen prefixes. The fixed-dose update acts only at the first decision; other tokens are omitted.}
\label{tab:single_step_control}
\label{tab:local_control}
\end{table}

\begin{table}[!htbp]
\centering\small
\setlength{\tabcolsep}{4pt}
\renewcommand{\arraystretch}{1.0}
\begin{tabular*}{\linewidth}{@{}@{\extracolsep{\fill}}lcc@{}}
\toprule
Method & Acc. (\%) & Tokens \\
\midrule
Original & 85.60 $\pm$ 1.00 & 1410.3 $\pm$ 11.8 \\
\rowcolor{oursgray}
Full & \textbf{87.13} $\pm$ 1.22 & 1281.5 $\pm$ 29.3 \\
NoGate/Floor & 85.60 $\pm$ 1.83 & \textbf{1150.2} $\pm$ 51.1 \\
NoAltGroup & 86.67 $\pm$ 1.27 & 1281.1 $\pm$ 42.6 \\
NoBoost & 85.67 $\pm$ 0.61 & 1326.3 $\pm$ 37.5 \\
NoRevSup & 86.33 $\pm$ 1.50 & 1330.7 $\pm$ 45.7 \\
\bottomrule
\end{tabular*}
\caption{Component ablation on MATH500: means and sample SD over three seeds. Full is gray; bold marks the best mean.}
\label{tab:math500_component_ablation}
\label{tab:historical_ablation_main}
\end{table}

\paragraph{Control settings choose different operating points.}
Base strength $\alpha_{\mathrm{base}}$ scales the update, while $\lambda_A$ weights alternative openings in the competing-marker score. Figure~\ref{fig:hyperparam_sensitivity} shows the accuracy--length changes that informed the defaults; Appendix Table~\ref{tab:historical_parameters} gives an independent repeated-run sweep.

Global Reverse lengthens generations on both AIME sets, while lowering mean accuracy on AIME2024 and raising it on AIME2025 relative to Forward. Adding \emph{Wait} preserves AIME2025 accuracy with shorter traces but lowers the AIME2024 mean. Direction, vocabulary, and generation-cap results appear in Appendix~\ref{app:rebuttal_supplements}.

Independent development comparisons likewise show shorter generations with mixed accuracy changes, including decreases on several Llama-8B tasks (Appendix~\ref{app:historical_tables}). Together, these findings identify task-dependent direction and strength as central choices in marker control.

\section{Conclusion}
Marker choice offers a useful handle for controlling reasoning, with effects that depend on both the marker and its context. PathCal turns this observation into a decoding rule that adjusts the strength of a fixed continuation bias using instantaneous marker competition. Its main evaluation shows improved accuracy with shorter generations in most settings, while additional tasks and ablations reveal where the tradeoff changes. These findings motivate further work on choosing intervention direction and strength for different reasoning tasks.

\section*{Limitations}
PathCal uses a fixed continuation direction and an English single-token marker inventory, making its accuracy--length tradeoff dependent on the task, model, and tokenizer; suppressing revision can also remove useful corrective steps. Small competition datasets and finite seed counts limit statistical precision. The coding and multilingual evaluations cover specific subsets: LiveCodeBench scores depend on the executed test cases, and MGSM scores depend on numeric-answer extraction across aligned translations of 50 base problems. The MGSM results primarily show shorter generations with similar accuracy means. Finally, generated-token reductions measure output length; practical latency and energy savings depend on implementation overhead and hardware.

\FloatBarrier
\bibliography{references}
\appendix
\newcommand{\apptablestyle}{\small\setlength{\tabcolsep}{5pt}\renewcommand{\arraystretch}{1.12}}
\newcommand{\apptabnote}[1]{\par\vspace{3pt}\begin{minipage}{\linewidth}\footnotesize\raggedright #1\end{minipage}}

\makeatletter
\setlength{\@fptop}{0pt}
\setlength{\@fpsep}{14pt}
\setlength{\@fpbot}{0pt plus 1fil}
\setlength{\@dblfptop}{0pt}
\setlength{\@dblfpsep}{14pt}
\setlength{\@dblfpbot}{0pt plus 1fil}
\makeatother

\section{Marker Vocabulary and Tokenizer Resolution}
\label{app:markers}

\textsc{PathCal} uses the 12 case-sensitive surface strings in
Table~\ref{tab:app_marker_vocab}. They define operational control groups:
the controller does not classify the semantic function of each occurrence.
For example, a token realizing ``But'' is assigned to the revision group
even when its use in context does not revise a reasoning step.

\begin{table*}[!tbp]
\centering
\caption{Canonical inventory of 12 case-sensitive marker strings.}
\label{tab:app_marker_vocab}
\apptablestyle
\begin{tabularx}{\linewidth}{@{}>{\raggedright\arraybackslash}p{3.7cm}>{\raggedright\arraybackslash}p{2.2cm}>{\raggedright\arraybackslash}X@{}}
\toprule
Category & Logit operation & Surface forms \\
\midrule
Continuation ($\mathcal{M}_C$) & Boost & \texttt{So}, \texttt{so}, \texttt{Therefore}, \texttt{therefore}, \texttt{Thus} \\
Revision ($\mathcal{M}_R$) & Suppress & \texttt{But}, \texttt{but}, \texttt{However}, \texttt{however}, \texttt{no} \\
Alternative opening ($\mathcal{M}_A$) & Suppress & \texttt{Alternatively}, \texttt{alternatively} \\
\bottomrule
\end{tabularx}
\apptabnote{Operations apply when the gate activates; categories do not verify the semantic role of every occurrence. Bare and leading-space realizations are retained only if they resolve to a single non-unknown token. Revision IDs from lower-case \texttt{but}/\texttt{no}, including retained leading-space variants, receive weight $w_v=1.5$; other revision IDs receive $1.0$. Diagnostic candidates \texttt{Wait}, \texttt{wait}, \texttt{Hmm}, and \texttt{hmm} are excluded from the controller.}
\end{table*}

\paragraph{Canonical and diagnostic inventories.}
The diagnostic inventory additionally considers \texttt{wait},
\texttt{Wait}, \texttt{hmm}, and \texttt{Hmm}. These four strings are
absent from the canonical controller. The diagnostic inventory therefore contains 16 strings, while the
controller uses 12. The number of resolved token IDs additionally
depends on the tokenizer. Appendix~\ref{app:rebuttal_supplements}
examines how expanding or reducing this inventory affects the
accuracy--length tradeoff.

\paragraph{Resolution procedure.}
For each surface string $s$, the implementation encodes both $s$ and
the variant formed by prepending one space, using
\texttt{tokenizer.encode(..., add\_special\_tokens=False)}.
It retains a realization only when it contains exactly one token and
its ID is not the unknown-token ID. Each class is the deduplicated
union of these IDs. The implementation requires the three
classes to be disjoint. Revision IDs obtained from lower-case
\texttt{but} or \texttt{no}, including their leading-space variants,
receive weight 1.5; other revision IDs receive weight 1.0.
The resulting groups provide a lexical interface for the decoding
controller.

The resulting token-ID sets depend on the tokenizer. Multi-token
realizations are omitted, and a word with a single-token realization may
still appear in other tokenization contexts. The inventory consists
of English markers; the MGSM experiments evaluate task performance
with this same inventory.

\FloatBarrier
\section{Canonical Configuration and Integration}
\label{app:algorithm}

Algorithm~\ref{alg:pathcal_decoding} in Section~\ref{sec:pathcal}
summarizes the decoding procedure. Table~\ref{tab:app_pathcal_hparams} lists
the canonical defaults; block-specific experiment settings appear in
Appendix~\ref{app:evidence}.

\begin{table*}[!tbp]
\centering
\caption{Canonical \textsc{PathCal} parameters.}
\label{tab:app_pathcal_hparams}
\apptablestyle
\begin{tabularx}{\linewidth}{@{}l>{\raggedright\arraybackslash}Xr@{}}
\toprule
Parameter & Meaning & Default \\
\midrule
$\alpha_{\mathrm{base}}$ & Maximum global intervention strength & $6.0$ \\
$\gamma$ & Margin in $B_t-C_t+\gamma$ & $0.05$ \\
$\tau$ & Gap saturation scale & $0.2$ \\
$\lambda_A$ & Alternative-score weight in $B_t=R_t+\lambda_A A_t$ & $1.5$ \\
$\beta_C$ & Continuation boost coefficient & $0.5$ \\
$\beta_R$ & Revision suppression coefficient & $1.0$ \\
$\beta_A$ & Alternative-opening suppression coefficient & $1.0$ \\
$\rho$ & Marker-score floor: skip when $C_t+B_t<\rho$ & $0.05$ \\
$\epsilon$ & Offset in the gate denominator & $10^{-3}$ \\
$w_v$ & Per-token revision weight & $1.0$ or $1.5$ \\
$\mathrm{minp}$ & Generated-token warmup before intervention & $100$ \\
\bottomrule
\end{tabularx}
\apptabnote{Revision IDs from lower-case \texttt{but}/\texttt{no} and their retained leading-space variants use $w_v=1.5$; other revision IDs use $1.0$. The saturation scale $\tau$ is distinct from sampling temperature. Block-specific vocabularies, ablations, fixed-dose probes, and generation settings are documented separately.}
\end{table*}

\paragraph{Gate interpretation.}
For $S_t=C_t+B_t>0$, the gate factors as
\begin{equation}
g_t=
\underbrace{\frac{4C_tB_t}{S_t^2}}_{\text{score balance}}
\underbrace{\frac{S_t^2}{S_t^2+\epsilon}}_{\text{score-strength attenuation}}.
\label{eq:pathcal_gate_factorization}
\end{equation}
Both factors lie in $[0,1]$. At a fixed total score, balance is
maximal when $C_t=B_t$; the second factor suppresses small total
scores. The explicit floor additionally skips low-score steps.
The gap factor in Eq.~\eqref{eq:pathcal_alpha} determines whether
competing-marker evidence is sufficiently close to, or above,
continuation evidence. Thus activation depends on the balance and
magnitude of the observable marker scores.

\paragraph{Reasoning-region integration.}
The processor receives generated token IDs and next-token logits.
It skips a step when the generated prefix is shorter than
$\mathrm{minp}$ or already contains the \texttt{</think>} token.
The check counts generated tokens, excluding the input prompt.
It does not independently search for an opening \texttt{<think>};
the prompt and model interface are expected to begin generation in
the reasoning region. The implementation requires \texttt{</think>}
to resolve to one token. These conditions define the reasoning region
in which the processor operates.

\paragraph{Direction control.}
The implementation also permits a diagnostic Reverse mode.
At the same supplied logits it computes the same scores, gate, and
nonnegative $\alpha_t$, then multiplies all marker shifts by $-1$.
This changes the intervention direction without changing its local
activation criterion. The direction is fixed for each condition.
Once generated prefixes diverge, subsequent scores and activations
can also diverge.

\FloatBarrier
\section{Local Log-Odds Property}
\label{app:logodds}
\label{app:pathcal_property_proof}

The property in Section~\ref{subsec:pathcal_property} describes the
effect of the additive update at a fixed decoding step. Let
$\Delta_t(v)=\tilde\ell_t(v)-\ell_t(v)$ and define
$p_t=\operatorname{softmax}(\ell_t)$ and
$q_t=\operatorname{softmax}(\tilde\ell_t)$.
For two tokens $u,v$ with finite compared logits, the respective
normalizers cancel:
\begin{equation}
\log\frac{q_t(u)}{q_t(v)}-
\log\frac{p_t(u)}{p_t(v)}
=\Delta_t(u)-\Delta_t(v).
\label{eq:pathcal_logodds_identity}
\end{equation}
The disjoint marker groups give
$\Delta_t(c)=\beta_C\alpha_t$ for $c\in\mathcal M_C$,
$\Delta_t(r)=-\beta_Rw_r\alpha_t$ for $r\in\mathcal M_R$,
and $\Delta_t(a)=-\beta_A\alpha_t$ for $a\in\mathcal M_A$.
Substitution gives Eq.~\eqref{eq:pathcal_revision_odds} and
\begin{equation}
\log\frac{q_t(c)}{q_t(a)}-
\log\frac{p_t(c)}{p_t(a)}
=(\beta_C+\beta_A)\alpha_t.
\label{eq:pathcal_alternative_odds}
\end{equation}
For positive coefficients and $\alpha_t>0$, the canonical Forward
rule strictly increases these continuation-to-competing-marker odds.
The update is finite and does not mask tokens; its effect need not be small. Since $\alpha_t\leq6$ and $w_r\leq1.5$ under the defaults, the continuation/revision log-odds increment is at most 12 before temperature, or 20 at $T=0.6$, before truncation. This identity characterizes the change in relative next-token probabilities at a fixed prefix. Section~\ref{sec:experiments} evaluates the resulting task-level accuracy and generation length.

\paragraph{Temperature and truncation.}
The controller's score computation uses the softmax of its supplied
logits, without an additional temperature parameter in that computation.
If a sampler subsequently divides both compared logit vectors by a
common temperature $T>0$, the corresponding pre-truncation log-odds
increment is $(\Delta_t(u)-\Delta_t(v))/T$.
If top-$p$ or top-$k$ truncation is subsequently applied, this ratio
identity remains meaningful only for pairs retained in both compared
supports; a removed token has zero sampling probability.
The vLLM 0.7.2 V0 and 0.19.0 V1 integrations apply PathCal before
temperature scaling and top-$p$ truncation.

\paragraph{Non-marker tokens.}
For a non-marker token, $\Delta_t(v)=0$. Its logit is unchanged,
but its normalized probability can change because marker shifts
alter the softmax normalizer. The controller introduces no hard mask, although the base sampler's
truncation can still remove tokens from the sampling support.

\section{Computational Cost}
\label{app:cost}

\textsc{PathCal} requires no parameter updates, extra model forward
pass per decoding step, external verifier, or additional sampled
candidate. It changes the logits produced during the existing rollout.

Let $K=|\mathcal M_C|+|\mathcal M_R|+|\mathcal M_A|$.
The canonical implementation explicitly casts the supplied logits
to float32 and computes a full-vocabulary softmax. This normalization
costs $O(|\mathcal V|)$; marker-score aggregation and logit edits cost
$O(K)$. The marker sets and scalar state require $O(K)$
storage, while the normalization uses an $O(|\mathcal V|)$ temporary
probability vector. A backend that already exposes the required
probabilities could avoid recomputing them, but that optimization is
not assumed here.

The Python processor also checks the generated history for
\texttt{</think>} and converts individual tensor values to scalars.
These operations can add backend-dependent overhead, including
device synchronization. Generated-token counts measure output length;
latency, throughput, and energy efficiency additionally depend on
the implementation and hardware.

\section{Evaluation Protocols}
\label{app:evidence}
The evaluation combines a main task comparison, independent development and component studies, coding and multilingual evaluations, and diagnostic interventions. Each study compares methods within its own protocol and baseline. The main comparison uses seed 42; the repeated development comparisons use three seeds; coding and multilingual mathematics use five seeds; direction and vocabulary studies use ten seeds. Differences in prompts, generation caps, tensor parallelism, and marker mappings are specified below and in Tables~\ref{tab:app_inference_config} and~\ref{tab:app_diagnostic_config}.

Three-seed comparisons report means and sample standard deviations across seeds. Five-seed and ten-seed comparisons report mean task scores and token counts. Fixed-prefix confidence intervals resample problem clusters, preserving their associated prefixes and rollouts. MGSM uses fifty base problems translated into eleven languages, so the translations are aligned observations rather than independent base problems.

Table~\ref{tab:app_datasets} lists dataset sizes and scoring rules.

\begin{table*}[!tbp]
\centering
\caption{Inference settings for independent development comparisons.}
\label{tab:app_inference_config}
\apptablestyle
\begin{tabular*}{\linewidth}{@{\extracolsep{\fill}}lrrrlr@{}}
\toprule
Model & Context & Token cap & TP & Precision & Comparison seeds \\
\midrule
Qwen-7B & 16384 & 8192 & 2 & float16 & 42, 123, 456 \\
Llama-8B & 16384 & 8192 & 4 & float16 & 42, 123, 456 \\
QwQ-32B & 40960 & 32768 & 4 & bfloat16 & 42 \\
\bottomrule
\end{tabular*}
\apptabnote{All rows use temperature $0.6$ and top-$p=0.95$. Qwen/Llama abbreviate DeepSeek-R1-Distill model names; TP is tensor parallelism. Context size and generated-token cap are distinct. The smaller distill-model sweeps disable top-$k$ and use a custom prompt ending in a reasoning opener. The larger QwQ block uses top-$k=40$ and its chat template; its direct comparisons use seed 42 only. The main evaluation uses the protocol in Appendix~\ref{app:evidence}.}
\end{table*}

\begin{table*}[!tbp]
\centering
\caption{Inference settings for Qwen-7B diagnostics and component ablations.}
\label{tab:app_diagnostic_config}
\apptablestyle
\begin{tabularx}{\linewidth}{@{}>{\raggedright\arraybackslash}p{4.0cm}rr>{\raggedright\arraybackslash}X@{}}
\toprule
Experiment & Context & Token cap & Scope \\
\midrule
MATH500 component ablation & 16384 & 8192 & TP4; independent baseline from the TP2 development sweep. \\
8192-token prefix study & 16384 & 8192 & MATH500: 1462 prefixes / 499 problems; AIME24: 90 / 30. \\
16384-token prefix study & 32768 & 16384 & AIME24 and AIME25: 90 prefixes / 30 problems each. \\
First-token dose probe & 16384 & 4096 & TP1, seed 9042; 120 prefixes / 60 problems; one-decision intervention. \\
\bottomrule
\end{tabularx}
\apptabnote{All rows use Qwen-7B, temperature $0.6$, and top-$p=0.95$. The gate-alignment probe reuses the first-token prefix bank with seed 42, TP1, and a 4096-token cap; it is not an independent prefix sample.}
\end{table*}

\begin{table*}[!tbp]
\centering
\caption{Evaluated datasets and denominators.}
\label{tab:app_datasets}
\apptablestyle
\begin{tabularx}{\linewidth}{@{}l>{\raggedright\arraybackslash}p{4.2cm}r>{\raggedright\arraybackslash}X@{}}
\toprule
Dataset & Task & $n$ & Scoring \\
\midrule
GSM8K & Grade-school mathematics & 1319 & Numeric answers \\
MATH500 & Competition mathematics & 500 & Numeric / expression answers \\
AMC2023 & Contest mathematics & 40 & Numeric answers \\
AIME2024 & Contest mathematics & 30 & Integer answers \\
AIME2025 & Contest mathematics & 30 & Integer answers \\
TheoremQA & Theorem-based reasoning & 800 & Numeric / short answers \\
\midrule
LiveCodeBench & Code generation & 279 & Local test-case execution (pass@1) \\
MGSM & Multilingual grade-school mathematics & $50\times11$ & Numeric-answer accuracy \\
\bottomrule
\end{tabularx}
\apptabnote{LiveCodeBench uses the v5 subset. MGSM contains fifty aligned base problems in eleven languages; translations and prefix rollouts do not add independent base problems. Development and prefix-collection scopes are specified in Appendix~\ref{app:development_protocol} and Table~\ref{tab:app_diagnostic_config}; coding and multilingual summaries appear in Table~\ref{tab:domain_summary}.}
\end{table*}

\paragraph{Main results.}
Table~\ref{tab:main_grouped} evaluates single-sample decoding with seed 42, temperature 0.6, and top-$p$ 0.95. Generation caps are 8192 tokens for the distill models and 16384 for QwQ long-budget AIME runs. AIME2024 and AIME2025 each contain 30 problems.

\subsection{Development and Component Protocols}
\label{app:development_protocol}
The supplementary comparisons in Appendix~\ref{app:historical_tables} use a separate protocol from the main comparison and robustness studies. Qwen-7B and Llama-8B use seeds 42, 123, and 456; the matched QwQ-32B comparison uses seed 42. Inference settings appear in Table~\ref{tab:app_inference_config}. The task sets comprise MATH500, AIME2024, AIME2025, AMC2023, GSM8K, and TheoremQA \citep{hendrycks2021math,cobbe2021gsm8k,chen-etal-2023-theoremqa,aimo2024amc}, with denominators in Table~\ref{tab:app_datasets}. MATH500 follows the process-supervision evaluation subset \citep{lightman2024lets}. MathArena highlights competition freshness and contamination as considerations in evaluating reasoning models \citep{balunović2026matharenaevaluatingllmsuncontaminated}.

The development controller uses $\alpha_{\mathrm{base}}=6$ and a 100-generated-token warmup. Accuracy and length are aggregated over the same seeds, with sample SD ($\mathrm{ddof}=1$) for multi-seed comparisons. In the experiment records, \texttt{SA-MC-WMBC@a6} identifies this controller and \texttt{PathCal-Full} identifies the independent component study. The supplementary CyclicReflex comparisons use a bare-form token mapping; their column label is Cyclic.

\paragraph{Component definitions.}
The MATH500 component study uses TP4 and its own Original baseline. NoGate/Floor sets the competition gate to one and bypasses the marker-score floor, while retaining the positive-gap condition, saturation, warmup, and end-token check. NoAltGroup sets both $\lambda_A=0$ and $\beta_A=0$, removing alternative scores and updates. The parameter row $\lambda_A=0$ changes only the score coefficient and still suppresses alternative IDs, so it is not identical to NoAltGroup. NoBoost sets $\beta_C=0$; NoRevSup disables the revision update while retaining the scoring rule.

\section{Additional Diagnostic Details}
\label{app:diagnostic_details}
\paragraph{Prefix probes.}
The 8192-token study contains 1462 MATH500 prefixes from 499 problems and 90 AIME2024 prefixes from 30 problems, with context 16384; the AIME2025 collection was incomplete. The 16384-token study contains 90 prefixes each for AIME2024 and AIME2025, with context 32768. Each prefix retains success counts from eight natural continuations and four per forced marker. Eight natural rollouts estimate $\widehat V(s)$, defining low ($\widehat V\leq.25$), middle ($.25<\widehat V<.75$), and high ($\widehat V\geq.75$) bins.

Tables~\ref{tab:prefix_all_ci} and~\ref{tab:prefix_16k_ci} give all bins. The negative middle-bin contrast in the shorter AIME2024 collection uses eight problem clusters and has substantial truncation. The collections are analyzed separately because their protocols differ. Bin assignments are rollout estimates, with finite-sample variation and endpoint floor/ceiling effects; the intervals account for correlated prefixes through problem-level clustering.

\begin{table*}[!tbp]
\caption{Fixed-prefix results with an 8192-token cap.}
\label{tab:prefix_all_ci}
\centering\apptablestyle
\begin{tabular*}{\linewidth}{@{\extracolsep{\fill}}llrrrrrl@{}}
\toprule
Dataset & Bin & Prefixes & Problems & So (\%) & But (\%) & $\Delta$ (pp) & 95\% CI \\
\midrule
MATH500 & low & 127 & 52 & 7.1 & 10.0 & $-3.0$ & $[-5.8,-0.2]$ \\
MATH500 & mid & 77 & 50 & 57.5 & 50.3 & $+7.1$ & $[-1.5,+15.2]$ \\
MATH500 & high & 1258 & 443 & 97.6 & 97.7 & $-0.1$ & $[-0.7,+0.4]$ \\
\addlinespace[3pt]
AIME2024 & low & 47 & 20 & 9.0 & 11.7 & $-2.7$ & $[-8.8,+3.8]$ \\
AIME2024 & mid & 12 & 8 & 41.7 & 56.2 & $-14.6$ & $[-22.9,-8.3]$ \\
AIME2024 & high & 31 & 12 & 90.3 & 88.7 & $+1.6$ & $[-1.7,+5.6]$ \\
\bottomrule
\end{tabular*}
\apptabnote{$\Delta=V_{\mathrm{So}}-V_{\mathrm{But}}$. Eight natural continuations determine each bin; four continuations evaluate each forced marker. Percentile intervals use 20,000 problem-cluster resamples (RNG 20261012), preserving all within-problem prefixes and success counts. They condition on observed bins and are not adjusted for multiple comparisons. Each generation-cap study is analyzed separately. The AIME2024 middle bin has only eight problem clusters and substantial truncation.}
\end{table*}

\begin{table*}[!tbp]
\caption{Fixed-prefix results with a 16384-token cap.}
\label{tab:prefix_16k_ci}
\centering\apptablestyle
\begin{tabular*}{\linewidth}{@{\extracolsep{\fill}}llrrrrrl@{}}
\toprule
Dataset & Bin & Prefixes & Problems & So (\%) & But (\%) & $\Delta$ (pp) & 95\% CI \\
\midrule
AIME2024 & low & 31 & 13 & 4.8 & 10.5 & $-5.6$ & $[-13.8,+0.0]$ \\
AIME2024 & mid & 20 & 12 & 58.8 & 52.5 & $+6.2$ & $[-2.5,+15.3]$ \\
AIME2024 & high & 39 & 16 & 89.7 & 89.1 & $+0.6$ & $[-5.0,+5.5]$ \\
\addlinespace[3pt]
AIME2025 & low & 46 & 18 & 8.2 & 7.1 & $+1.1$ & $[-2.9,+5.4]$ \\
AIME2025 & mid & 17 & 9 & 36.8 & 47.1 & $-10.3$ & $[-25.0,+7.5]$ \\
AIME2025 & high & 27 & 10 & 95.4 & 96.3 & $-0.9$ & $[-8.3,+6.2]$ \\
\bottomrule
\end{tabular*}
\apptabnote{$\Delta=V_{\mathrm{So}}-V_{\mathrm{But}}$. Eight natural continuations determine each bin; four continuations evaluate each forced marker. Percentile intervals use 20,000 problem-cluster resamples (RNG 20261012), preserving all within-problem prefixes and success counts. They condition on observed bins and are not adjusted for multiple comparisons. Each generation-cap study is analyzed separately.}
\end{table*}

\paragraph{Local first-token control.}
The first-token experiment uses a fixed Qwen-7B model revision, seed 9042, temperature 0.6, top-$p$ 0.95, cap 4096, context 16384, TP1, and 120 prefixes from 60 problems. Its intervention applies only at the first decision. Each condition has four rollouts per prefix. The direction score is
\begin{equation}
S=\tfrac12[(f_C^{F}-f_C^{R})+(f_{R\cup A}^{R}-f_{R\cup A}^{F})],
\end{equation}
where $f$ denotes first-token frequency and $F/R$ denotes Forward/Reverse at strength 6. Confidence intervals resample the 60 problem clusters 20000 times (RNG 20260802), retaining both prefixes and their rollouts. The real/placebo contrast is paired by problem. Placebo frequency matching uses global token frequencies rather than prefix-specific probability mass.

Table~\ref{tab:ordinary_token_control} reports the marker/placebo comparison.

\begin{table}[!tbp]
\caption{Marker versus ordinary-token control.}
\label{tab:ordinary_token_control}
\centering\apptablestyle
\begin{tabular*}{\linewidth}{@{\extracolsep{\fill}}lrr@{}}
\toprule
Inventory & Mean $S$ & 95\% CI \\
\midrule
Markers & 0.6281 & [0.5656, 0.6885] \\
Ordinary tokens & 0.0292 & [0.0135, 0.0469] \\
\midrule
Difference & 0.5990 & [0.5354, 0.6625] \\
\bottomrule
\end{tabular*}
\apptabnote{$S$ is dimensionless signed category displacement, not accuracy. The paired difference uses 20,000 resamples over 60 problem clusters (RNG 20260802). Ordinary tokens are globally frequency-matched, not matched on per-prefix probability mass.}
\end{table}

\paragraph{Gate alignment probe.}
This probe uses the same Qwen-7B AIME24/25 prefix bank as the first-token control study with 60 prefixes per dataset, seed 42, cap 4096, and TP1. Every prefix has four normal, continuation, and revision rollouts. The recorded gate is $g_t$, set to zero below the marker-score floor; a prefix is triggered when $\alpha_t>0$ under the canonical configuration. Four rollouts per direction give a success-rate resolution of 0.25, so ties need not mean identical underlying values. Table~\ref{tab:gate_alignment} reports descriptive prefix-level correlations and direction counts among triggered prefixes.

\begin{table}[!tbp]
\caption{Gate alignment on AIME2024/2025 prefixes.}
\label{tab:gate_alignment}
\centering\apptablestyle
\begin{tabular*}{\linewidth}{@{\extracolsep{\fill}}lr@{}}
\toprule
Measure & Value \\
\midrule
Prefixes & 120 \\
Problems & 60 \\
Triggered prefixes & 40 \\
\midrule
$\rho(g,\mathrm{sensitivity})$ & $-0.111$ \\
$\rho(g,\mathrm{midness})$ & $-0.142$ \\
\midrule
Continuation better & 3 \\
Revision better & 4 \\
Tied & 33 \\
\bottomrule
\end{tabular*}
\apptabnote{Spearman correlations use all prefixes. Sensitivity is the absolute forced-direction success difference; midness is $\max(1-2|\widehat V-0.5|,0)$. Direction counts use the 40 triggered prefixes. Problems are the independent task units. These descriptive associations do not support positive semantic-state alignment.}
\end{table}

\section{Supplementary Comparisons and Ablations}
\label{app:historical_tables}
The independent development comparisons test whether the accuracy--length tradeoff persists across repeated evaluations and additional tasks. PathCal shortens generations for both Qwen-7B and Llama-8B on all six tasks in Tables~\ref{tab:historical_main} and~\ref{tab:historical_main_tokens}, while accuracy changes depend on the model and task. Both models improve on AIME2024; Llama-8B has lower accuracy on the other five tasks. Thus the reduction in generation length is more consistent than the accuracy benefit in this study. Protocols and aggregation are specified in Appendix~\ref{app:development_protocol}.

\begin{table*}[!tbp]
\centering
\apptablestyle
\caption{Three-seed development comparison: accuracy (\%).}
\label{tab:historical_main}
\begin{tabular*}{\textwidth}{@{\extracolsep{\fill}}lrrrr@{}}
\toprule
Dataset & $n$ & Original & PathCal & $\Delta$ (pp) \\
\midrule
\multicolumn{5}{@{}l}{\textit{Qwen-7B}} \\
MATH500 & 500 & 85.47 $\pm$ 0.64 & 86.27 $\pm$ 1.21 & +0.80 \\
AIME2024 & 30 & 32.22 $\pm$ 6.94 & 45.56 $\pm$ 10.18 & +13.33 \\
AIME2025 & 30 & 26.67 $\pm$ 3.33 & 31.11 $\pm$ 3.85 & +4.44 \\
AMC2023 & 40 & 77.50 $\pm$ 6.61 & 76.67 $\pm$ 7.22 & -0.83 \\
GSM8K & 1319 & 86.76 $\pm$ 0.22 & 86.63 $\pm$ 0.31 & -0.13 \\
TheoremQA & 800 & 43.21 $\pm$ 0.80 & 43.67 $\pm$ 0.73 & +0.46 \\
\midrule
\multicolumn{5}{@{}l}{\textit{Llama-8B}} \\
MATH500 & 500 & 83.13 $\pm$ 0.76 & 82.33 $\pm$ 0.81 & -0.80 \\
AIME2024 & 30 & 32.22 $\pm$ 1.92 & 40.00 $\pm$ 3.33 & +7.78 \\
AIME2025 & 30 & 27.78 $\pm$ 3.85 & 25.56 $\pm$ 5.09 & -2.22 \\
AMC2023 & 40 & 77.50 $\pm$ 2.50 & 72.50 $\pm$ 2.50 & -5.00 \\
GSM8K & 1319 & 84.05 $\pm$ 0.93 & 83.78 $\pm$ 0.87 & -0.28 \\
TheoremQA & 800 & 39.17 $\pm$ 1.05 & 37.79 $\pm$ 0.40 & -1.37 \\
\bottomrule
\end{tabular*}
\apptabnote{Means $\pm$ sample SD across seeds 42, 123, and 456 ($\mathrm{ddof}=1$). $\Delta$ is PathCal minus Original in percentage points. Both models use an 8192-token cap; Qwen-7B uses TP2 and Llama-8B TP4.}
\end{table*}

\begin{table*}[!tbp]
\centering
\apptablestyle
\caption{Three-seed development comparison: generated tokens.}
\label{tab:historical_main_tokens}
\begin{tabular*}{\textwidth}{@{\extracolsep{\fill}}lrrrr@{}}
\toprule
Dataset & $n$ & Original & PathCal & $\Delta$ (\%) \\
\midrule
\multicolumn{5}{@{}l}{\textit{Qwen-7B}} \\
MATH500 & 500 & 1423.0 $\pm$ 25.6 & 1291.6 $\pm$ 15.7 & -9.23\% \\
AIME2024 & 30 & 5678.4 $\pm$ 641.9 & 4917.5 $\pm$ 262.9 & -13.40\% \\
AIME2025 & 30 & 5606.0 $\pm$ 12.8 & 5244.6 $\pm$ 184.3 & -6.45\% \\
AMC2023 & 40 & 2749.7 $\pm$ 185.4 & 2443.3 $\pm$ 181.2 & -11.14\% \\
GSM8K & 1319 & 338.6 $\pm$ 14.6 & 330.3 $\pm$ 15.1 & -2.46\% \\
TheoremQA & 800 & 1640.0 $\pm$ 49.1 & 1423.0 $\pm$ 31.3 & -13.23\% \\
\midrule
\multicolumn{5}{@{}l}{\textit{Llama-8B}} \\
MATH500 & 500 & 1967.9 $\pm$ 74.1 & 1867.0 $\pm$ 4.7 & -5.13\% \\
AIME2024 & 30 & 6358.6 $\pm$ 257.1 & 6040.9 $\pm$ 355.4 & -5.00\% \\
AIME2025 & 30 & 6619.8 $\pm$ 308.5 & 5953.5 $\pm$ 215.4 & -10.07\% \\
AMC2023 & 40 & 3529.8 $\pm$ 394.7 & 3277.8 $\pm$ 194.0 & -7.14\% \\
GSM8K & 1319 & 474.6 $\pm$ 34.5 & 448.6 $\pm$ 34.2 & -5.47\% \\
TheoremQA & 800 & 2505.0 $\pm$ 60.6 & 2152.0 $\pm$ 24.9 & -14.09\% \\
\bottomrule
\end{tabular*}
\apptabnote{Means $\pm$ sample SD across seeds 42, 123, and 456 ($\mathrm{ddof}=1$). $\Delta$ is the relative change from Original; negative values indicate shorter traces. Both models use an 8192-token cap; Qwen-7B uses TP2 and Llama-8B TP4.}
\end{table*}

\begin{table*}[!tbp]
\centering
\apptablestyle
\caption{Llama-8B: three-seed matched comparison.}
\label{tab:historical_llama_threeway}
\begin{tabular*}{\textwidth}{@{\extracolsep{\fill}}lrrr@{}}
\toprule
Dataset & Original & PathCal & Cyclic \\
\midrule
\multicolumn{4}{@{}l}{\textit{(A) Accuracy (\%)}} \\
MATH500 & 83.13 $\pm$ 0.76 & 82.33 $\pm$ 0.81 & 83.20 $\pm$ 1.91 \\
AIME2024 & 32.22 $\pm$ 1.92 & 40.00 $\pm$ 3.33 & 38.89 $\pm$ 3.85 \\
AIME2025 & 27.78 $\pm$ 3.85 & 25.56 $\pm$ 5.09 & 27.78 $\pm$ 1.92 \\
AMC2023 & 77.50 $\pm$ 2.50 & 72.50 $\pm$ 2.50 & 81.67 $\pm$ 5.77 \\
GSM8K & 84.05 $\pm$ 0.93 & 83.78 $\pm$ 0.87 & 84.94 $\pm$ 1.81 \\
TheoremQA & 39.17 $\pm$ 1.05 & 37.79 $\pm$ 0.40 & 39.12 $\pm$ 1.27 \\
\midrule
\multicolumn{4}{@{}l}{\textit{(B) Generated tokens}} \\
MATH500 & 1967.9 $\pm$ 74.1 & 1867.0 $\pm$ 4.7 & 2017.5 $\pm$ 62.1 \\
AIME2024 & 6358.6 $\pm$ 257.1 & 6040.9 $\pm$ 355.4 & 6472.5 $\pm$ 134.0 \\
AIME2025 & 6619.8 $\pm$ 308.5 & 5953.5 $\pm$ 215.4 & 6054.7 $\pm$ 460.0 \\
AMC2023 & 3529.8 $\pm$ 394.7 & 3277.8 $\pm$ 194.0 & 3264.2 $\pm$ 204.5 \\
GSM8K & 474.6 $\pm$ 34.5 & 448.6 $\pm$ 34.2 & 524.6 $\pm$ 51.3 \\
TheoremQA & 2505.0 $\pm$ 60.6 & 2152.0 $\pm$ 24.9 & 2503.7 $\pm$ 58.0 \\
\bottomrule
\end{tabular*}
\apptabnote{Means $\pm$ sample SD across seeds 42, 123, and 456; TP4; 8192-token cap. Cyclic denotes CyclicReflex with bare-form token mapping (Appendix~\ref{app:development_protocol}).}
\end{table*}

\begin{table*}[!tbp]
\centering
\apptablestyle
\caption{QwQ-32B: matched seed-42 comparison.}
\label{tab:historical_qwq_seed42}
\begin{tabular*}{\textwidth}{@{\extracolsep{\fill}}lrrr@{}}
\toprule
Dataset & Original & PathCal & Cyclic \\
\midrule
\multicolumn{4}{@{}l}{\textit{(A) Accuracy (\%)}} \\
MATH500 & 94.80 & 93.20 & 93.20 \\
AIME2024 & 73.33 & 80.00 & 73.33 \\
AIME2025 & 66.67 & 70.00 & 76.67 \\
AMC2023 & 100.00 & 100.00 & 100.00 \\
GSM8K & 96.66 & 96.29 & 96.66 \\
TheoremQA & 58.50 & 59.25 & 58.88 \\
\midrule
\multicolumn{4}{@{}l}{\textit{(B) Generated tokens}} \\
MATH500 & 4209.4 & 3951.1 & 4261.3 \\
AIME2024 & 12885.8 & 13725.8 & 12848.3 \\
AIME2025 & 15792.4 & 16183.4 & 14733.3 \\
AMC2023 & 7524.4 & 6608.4 & 7076.4 \\
GSM8K & 1593.2 & 1437.9 & 1580.6 \\
TheoremQA & 6847.9 & 6006.9 & 6685.7 \\
\bottomrule
\end{tabular*}
\apptabnote{Seed 42; TP4; 32768-token cap; QwQ chat template. Cyclic denotes CyclicReflex with bare-form token mapping (Appendix~\ref{app:development_protocol}).}
\end{table*}

The matched comparisons with CyclicReflex further illustrate the tradeoff (Tables~\ref{tab:historical_llama_threeway} and~\ref{tab:historical_qwq_seed42}). CyclicReflex is both more accurate and shorter than PathCal on Llama-8B AMC2023 and QwQ-32B AIME2025. The relative advantage of a marker-control rule therefore depends on the evaluation setting.

\paragraph{Complete component and parameter sweep.}
Table~\ref{tab:historical_ablation} expands the component summary; Table~\ref{tab:historical_parameters} separately lists all tested strengths and alternative-marker weights.
\begin{table*}[!tbp]
\centering
\apptablestyle
\caption{Component ablation on MATH500.}
\label{tab:historical_ablation}
\begin{tabular*}{\textwidth}{@{\extracolsep{\fill}}lrrrrr@{}}
\toprule
Setting & Accuracy (\%) & Tokens & \shortstack{$\Delta$ acc.\\vs. Original} & \shortstack{$\Delta$ tokens\\vs. Original} & \shortstack{$\Delta$ acc.\\vs. Full} \\
\midrule
Original & 85.60 $\pm$ 1.00 & 1410.3 $\pm$ 11.8 & +0.00 & +0.00\% & -1.53 \\
Full & 87.13 $\pm$ 1.22 & 1281.5 $\pm$ 29.3 & +1.53 & -9.13\% & +0.00 \\
NoGate/Floor & 85.60 $\pm$ 1.83 & 1150.2 $\pm$ 51.1 & +0.00 & -18.44\% & -1.53 \\
NoAltGroup & 86.67 $\pm$ 1.27 & 1281.1 $\pm$ 42.6 & +1.07 & -9.16\% & -0.47 \\
NoBoost & 85.67 $\pm$ 0.61 & 1326.3 $\pm$ 37.5 & +0.07 & -5.96\% & -1.47 \\
NoRevSup & 86.33 $\pm$ 1.50 & 1330.7 $\pm$ 45.7 & +0.73 & -5.64\% & -0.80 \\
\bottomrule
\end{tabular*}
\apptabnote{Qwen-7B on MATH500 (500 problems); TP4; 8192-token cap; seeds 42, 123, and 456. Means $\pm$ sample SD ($\mathrm{ddof}=1$). Accuracy differences are percentage points; token differences are relative percentages. All differences use this study's own Original or Full row. NoGate/Floor disables the competition gate and bypasses the marker-score floor; NoAltGroup removes the alternative class; NoBoost and NoRevSup disable the named update. See Appendix~\ref{app:development_protocol} for the exact definitions.}
\end{table*}

\begin{table*}[!tbp]
\centering
\apptablestyle
\caption{Parameter sweep on MATH500.}
\label{tab:historical_parameters}
\begin{tabular*}{\textwidth}{@{\extracolsep{\fill}}lrrrrr@{}}
\toprule
Setting & Accuracy (\%) & Tokens & \shortstack{$\Delta$ acc.\\vs. Original} & \shortstack{$\Delta$ tokens\\vs. Original} & \shortstack{$\Delta$ acc.\\vs. Full} \\
\midrule
Full (reference) & 87.13 $\pm$ 1.22 & 1281.5 $\pm$ 29.3 & +1.53 & -9.13\% & +0.00 \\
\midrule
\multicolumn{6}{@{}l}{\textit{Strength $\alpha_{\mathrm{base}}$}} \\
$\alpha_{\mathrm{base}}=2$ & 86.47 $\pm$ 1.42 & 1325.9 $\pm$ 26.7 & +0.87 & -5.98\% & -0.67 \\
$\alpha_{\mathrm{base}}=4$ & 86.13 $\pm$ 0.12 & 1279.9 $\pm$ 16.2 & +0.53 & -9.25\% & -1.00 \\
$\alpha_{\mathrm{base}}=8$ & 86.80 $\pm$ 0.92 & 1288.2 $\pm$ 12.2 & +1.20 & -8.66\% & -0.33 \\
$\alpha_{\mathrm{base}}=10$ & 86.27 $\pm$ 1.17 & 1243.4 $\pm$ 13.7 & +0.67 & -11.84\% & -0.87 \\
\midrule
\multicolumn{6}{@{}l}{\textit{Alternative-marker weight $\lambda_A$}} \\
$\lambda_A=0$ & 86.20 $\pm$ 1.97 & 1320.0 $\pm$ 38.1 & +0.60 & -6.40\% & -0.93 \\
$\lambda_A=0.5$ & 86.73 $\pm$ 0.31 & 1309.8 $\pm$ 12.0 & +1.13 & -7.13\% & -0.40 \\
$\lambda_A=1$ & 87.00 $\pm$ 0.69 & 1293.5 $\pm$ 39.8 & +1.40 & -8.29\% & -0.13 \\
$\lambda_A=2$ & 87.07 $\pm$ 0.99 & 1273.0 $\pm$ 18.8 & +1.47 & -9.73\% & -0.07 \\
\bottomrule
\end{tabular*}
\apptabnote{Qwen-7B on MATH500 (500 problems); TP4; 8192-token cap; seeds 42, 123, and 456. Means $\pm$ sample SD ($\mathrm{ddof}=1$). Accuracy differences are percentage points; token differences are relative percentages. All differences use this study's own Original or Full row. Full is repeated as a reference. Each parameter row changes only the named setting; $\lambda_A=0$ is distinct from NoAltGroup. The sweep measures sensitivity to individual parameters; settings were not selected on a held-out evaluation.}
\end{table*}

\section{Additional Robustness and Scope Analyses}
\label{app:rebuttal_supplements}
\paragraph{Fixed-direction control.}
To isolate the role of intervention direction, the ten-seed comparison in Table~\ref{tab:rebuttal_reverse} reverses the logit adjustments while keeping the activation rule fixed. Reverse is longer on both tasks, with lower accuracy on AIME2024 and higher accuracy on AIME2025 relative to Forward. Forward therefore produces shorter reasoning traces in these comparisons, while the preferred direction for accuracy varies by task. Each direction is fixed throughout generation.
\begin{table*}[!tbp]
\centering\apptablestyle
\caption{Ten-seed global direction control on Qwen-7B.}
\label{tab:rebuttal_reverse}
\begin{tabular*}{\textwidth}{@{\extracolsep{\fill}}llrr@{}}
\toprule
Dataset & Method & Accuracy (\%) & Tokens \\
\midrule
AIME2024 & Original & 35.33 & 5615.9 \\
& Forward & 46.67 & 4976.0 \\
& Reverse & 43.33 & 5797.8 \\
\midrule
AIME2025 & Original & 26.67 & 5747.3 \\
& Forward & 30.00 & 5078.3 \\
& Reverse & 32.67 & 5969.8 \\
\bottomrule
\end{tabular*}
\apptabnote{Means over ten seeds. Forward is the PathCal direction; Reverse applies the opposite direction throughout generation under the same activation rule. The first-token probe in Table~\ref{tab:single_step_control} instead intervenes at a single decision.}
\end{table*}
\paragraph{Vocabulary sensitivity.}
The marker inventory determines which lexical transitions the controller can influence. Table~\ref{tab:rebuttal_vocabulary} compares five fixed inventories over ten seeds to examine this dependence. Canonical has the highest AIME2024 mean, whereas PlusWait matches its AIME2025 accuracy with shorter traces. No inventory dominates both tasks, indicating that adding reflection forms changes the tradeoff rather than providing a uniform improvement. This post-hoc sensitivity study uses manually specified inventories, without held-out vocabulary selection.
\begin{table*}[!tbp]
\centering\apptablestyle
\caption{Ten-seed vocabulary sensitivity on Qwen-7B.}
\label{tab:rebuttal_vocabulary}
\begin{tabular*}{\textwidth}{@{\extracolsep{\fill}}lrrrrr@{}}
\toprule
Variant & IDs (C/R/A) & \multicolumn{2}{c}{AIME2024} & \multicolumn{2}{c}{AIME2025} \\
\cmidrule(lr){3-4}\cmidrule(l){5-6}
& & Acc. (\%) & Tokens & Acc. (\%) & Tokens \\
\midrule
Canonical & 9/10/3 & 46.67 & 4976.0 & 30.00 & 5078.3 \\
Core & 4/4/3 & 43.33 & 5352.0 & 27.33 & 5309.6 \\
PlusWait & 9/14/3 & 37.33 & 5173.6 & 30.00 & 4562.0 \\
WaitOnly & 9/4/0 & 40.00 & 5020.3 & 22.00 & 5179.1 \\
Diagnostic-expanded & 9/17/3 & 40.67 & 5037.8 & 24.00 & 4696.7 \\
\bottomrule
\end{tabular*}
\apptabnote{ID counts refer to resolved continuation/revision/alternative tokens, not surface strings. Values are means over ten seeds.}
\end{table*}
\paragraph{Generation budget.}
Table~\ref{tab:rebuttal_caps} examines AIME2024 performance under three generation caps with seed 42. Its 8192-cap entries match the main comparison, with accuracy shown to two decimal places from the counts out of 30. PathCal achieves higher accuracy than Original at each cap and uses fewer tokens on average. At the full cap, PathCal is more accurate than TIP, while TIP generates fewer tokens. These comparisons illustrate distinct accuracy--length operating points under a shared maximum budget; CyclicReflex is outside the scope of this sweep.
\begin{table*}[!tbp]
\centering\apptablestyle
\caption{Partial generation-cap sweep on Qwen-7B AIME2024.}
\label{tab:rebuttal_caps}
\begin{tabular*}{\textwidth}{@{\extracolsep{\fill}}rlrr@{}}
\toprule
Token cap & Method & Accuracy (\%) & Tokens \\
\midrule
2048 & Original & 10.00 & 1969.3 \\
& PathCal & 13.33 & 1929.7 \\
& TIP & 10.00 & 1969.3 \\
& s1 & 10.00 & 2023.6 \\
\midrule
4096 & Original & 16.67 & 3716.1 \\
& PathCal & 30.00 & 3189.1 \\
& TIP & 20.00 & 3634.5 \\
& s1 & 30.00 & 3739.2 \\
\midrule
8192 & Original & 33.33 & 5678.0 \\
& PathCal & 43.33 & 4990.0 \\
& TIP & 36.67 & 4772.0 \\
& s1 & 40.00 & 6154.0 \\
\bottomrule
\end{tabular*}
\apptabnote{Seed 42. A shared generation cap does not imply matched realized token use; CyclicReflex is absent from this partial sweep.}
\end{table*}
\paragraph{Activation and marker emissions.}
On the logged Qwen-7B AIME2024/AIME2025 subsets, trace coverage is 68.7/65.3\%, and the controller triggers on 2.124/2.468\% of decoding steps. Continuation emissions change by +16/+1\%, revision emissions by $-30/-38\%$, and alternative emissions by $-23/-28\%$. At least one trigger occurs in 99.0/100.0\% of logged traces, with 74.89/81.28 mean triggers per trace. Relative trigger-position quantiles (10th/50th/90th) are 0.13/0.53/0.92 and 0.14/0.51/0.91. The controller thus acts at a small fraction of decoding steps but across nearly all logged traces, with triggers distributed throughout generation. These are activation and emission statistics for the logged subset. The separate prefix probe in Table~\ref{tab:gate_alignment} examines whether activation aligns with a benefit from continuation.

\section{Tokenizer Mappings}
\label{app:tokenizer_mapping}
Tables~\ref{tab:tokenizer_qwen} and~\ref{tab:tokenizer_llama} encode the twelve canonical strings and four excluded candidates for the five listed tokenizers. DeepSeek-R1-Distill-Qwen-32B is included only for tokenizer compatibility. Each canonical mapping has 9 continuation, 10 revision, and 3 alternative IDs. Multi-token encodings are excluded. Tokenizers with identical mappings share a table; the mappings specify the operational control vocabulary.

\begin{table*}[!t]
\centering
\caption{Tokenizer mappings for the four Qwen-family checkpoints.}
\label{tab:tokenizer_qwen}
\apptablestyle
\begin{tabular*}{\linewidth}{@{\extracolsep{\fill}}lcrr@{}}
\toprule
Surface form & Used & Bare token IDs & Leading-space token IDs \\
\midrule
\multicolumn{4}{@{}l}{\textit{Canonical controller: 12 strings}} \\
\texttt{So} & yes & [4416] & [2055] \\
\texttt{so} & yes & [704] & [773] \\
\texttt{Therefore} & yes & [54815] & [15277] \\
\texttt{therefore} & yes & [18532,1324] & [8916] \\
\texttt{Thus} & yes & [44500] & [14301] \\
\texttt{But} & yes & [3983] & [1988] \\
\texttt{but} & yes & [8088] & [714] \\
\texttt{However} & yes & [11209] & [4354] \\
\texttt{however} & yes & [97836] & [4764] \\
\texttt{no} & yes & [2152] & [902] \\
\texttt{Alternatively} & yes & [92014] & [38478] \\
\texttt{alternatively} & yes & [40412,7887] & [68387] \\
\midrule
\multicolumn{4}{@{}l}{\textit{Diagnostic candidates excluded from the controller: 4 strings}} \\
\texttt{Wait} & no & [14190] & [13824] \\
\texttt{wait} & no & [11489] & [3783] \\
\texttt{Hmm} & no & [80022] & [88190] \\
\texttt{hmm} & no & [71,3821] & [87501] \\
\bottomrule
\end{tabular*}
\apptabnote{Encodings apply to QwQ-32B, Qwen-14B, Distill-Qwen-32B, and Qwen-7B. \textit{Used} denotes membership in the canonical controller inventory. Bare and leading-space encodings are checked separately; bracketed multi-ID realizations are excluded.}
\end{table*}

\begin{table*}[!t]
\centering
\caption{Tokenizer mappings for Llama-8B.}
\label{tab:tokenizer_llama}
\apptablestyle
\begin{tabular*}{\linewidth}{@{\extracolsep{\fill}}lcrr@{}}
\toprule
Surface form & Used & Bare token IDs & Leading-space token IDs \\
\midrule
\multicolumn{4}{@{}l}{\textit{Canonical controller: 12 strings}} \\
\texttt{So} & yes & [4516] & [2100] \\
\texttt{so} & yes & [708] & [779] \\
\texttt{Therefore} & yes & [55915] & [15636] \\
\texttt{therefore} & yes & [19041,1348] & [9093] \\
\texttt{Thus} & yes & [45600] & [14636] \\
\texttt{But} & yes & [4071] & [2030] \\
\texttt{but} & yes & [8248] & [719] \\
\texttt{However} & yes & [11458] & [4452] \\
\texttt{however} & yes & [98936] & [4869] \\
\texttt{no} & yes & [2201] & [912] \\
\texttt{Alternatively} & yes & [93114] & [39578] \\
\texttt{alternatively} & yes & [41512,8046] & [69487] \\
\midrule
\multicolumn{4}{@{}l}{\textit{Diagnostic candidates excluded from the controller: 4 strings}} \\
\texttt{Wait} & no & [14524] & [14144] \\
\texttt{wait} & no & [11748] & [3868] \\
\texttt{Hmm} & no & [81122] & [89290] \\
\texttt{hmm} & no & [71,3906] & [88601] \\
\bottomrule
\end{tabular*}
\apptabnote{Encodings apply to Llama-8B. \textit{Used} denotes membership in the canonical controller inventory. Bare and leading-space encodings are checked separately; bracketed multi-ID realizations are excluded.}
\end{table*}

\end{document}